\documentclass[10pt,twocolumn,letterpaper]{article}

\usepackage{cvpr}              
\definecolor{cvprblue}{rgb}{0.21,0.49,0.74}
\usepackage[pagebackref,breaklinks,colorlinks,allcolors=cvprblue]{hyperref}
\usepackage{wrapfig}
\usepackage{amsmath}
\usepackage{graphicx}
\usepackage{subcaption}
\usepackage{url}
\usepackage{hyperref}
\usepackage{booktabs}
\usepackage{enumitem}
\usepackage{pifont}
\usepackage{multirow}
\usepackage{tcolorbox}
\usepackage{IEEEtrantools}
\usepackage{adjustbox}
\usepackage{threeparttable}
\usepackage{minted}
\usepackage{marvosym}
\usepackage{listings}

\def\confName{CVPR}
\def\confYear{2026}

\newcommand{\Dataset}{Recap-OmniCAD\textsuperscript{+}}

\newcommand{\commandnotation}[1]{\textless \textit{#1}\textgreater}
\newcommand{\sequencenotation}[1]{[\textit{#1}]}
\newcommand{\MethodName}{Pointer-CAD}
\newcommand{\cmark}{\ding{51}}  
\newcommand{\xmark}{\ding{55}}  

\title{Pointer-CAD: Unifying B-Rep and Command Sequences \\ via Pointer-based Edges \& Faces Selection}

\author{
Dacheng Qi$^{*,\dagger,1,2}$ \quad
Chenyu Wang$^{*,3,4}$ \quad
Jingwei Xu$^{\ddagger,5}$ \quad
Tianzhe Chu$^{3}$ \quad \\ 
Zibo Zhao$^{5}$ \quad
Wen Liu$^{6}$\quad
Wenrui Ding$^{2}$\quad
Yi Ma$^{1,3,4,7}$\quad
Shenghua Gao\textsuperscript{\Letter}$^{,1,3,4}$
\vspace{8pt}\\
	$^{1}$Transcengram\qquad
  $^{2}$Beihang University\qquad
  $^{3}$The University of Hong Kong\qquad  \\
  $^{4}$Shenzhen Loop Area Institute\qquad
  $^{5}$ShanghaiTech University\qquad \\
  $^{6}$DeepSeek\qquad
  $^{7}$University of California, Berkeley \\
}

\begin{document}
\maketitle
{
  \begingroup
  \renewcommand\thefootnote{}
  \footnotetext{* Equal contribution. \quad $\ddagger$ Also make major contribution.}
  \footnotetext{$\dagger$ Work done during an internship at Transcengram.}
  \footnotetext{\Letter\ Corresponding author: gaosh@hku.hk}
  \endgroup
}

\begin{abstract}
   Constructing computer-aided design (CAD) models is labor-intensive but essential for engineering and manufacturing. Recent advances in Large Language Models (LLMs) have inspired the LLM-based CAD generation by representing CAD as command sequences. But these methods struggle in practical scenarios because command sequence representation does not support entity selection~(e.g. faces or edges), limiting its ability to support complex editing operations such as \textit{chamfer} or \textit{fillet}. 
Further, the discretization of a continuous variable during \textit{sketch} and \textit{extrude} operations may result in topological errors.
To address these limitations, we present \textbf{Pointer-CAD}, a novel LLM-based CAD generation framework that leverages a pointer-based command sequence representation to explicitly incorporate the geometric information of B-rep models into sequential modeling. In particular, Pointer-CAD decomposes CAD model generation into steps, conditioning the generation of each subsequent step on both the textual description and the B-rep generated from previous steps. Whenever an operation requires the selection of a specific geometric entity, the LLM predicts a \textit{Pointer} that selects the most feature-consistent candidate from the available set. Such a selection operation also reduces the quantization error in the command sequence-based representation.
To support the training of Pointer-CAD, we develop a data annotation pipeline that produces expert-level natural language descriptions and apply it to build a dataset of approximately 575K CAD models.
Extensive experimental results demonstrate that Pointer-CAD effectively supports the generation of complex geometric structures and reduces segmentation error to an extremely low level, achieving a significant improvement over prior command sequence methods, thereby significantly mitigating the topological inaccuracies introduced by quantization error.
Our code is available at \url{https://github.com/Snitro/Pointer-CAD}.

\end{abstract}

\section{Introduction}
Computer-Aided Design (CAD) plays an essential role in modern engineering, enabling precise and efficient design across diverse industry domains ~\cite{rapp2021mlcad,castellino2005computer}. The conventional CAD design workflow typically begins with 2D sketches(e.g. \textit{lines, circles}), progresses to 3D modeling operations(e.g. \textit{extrude, chamfer, fillet}), and culminates in models stored in Boundary Representation (B-rep) \citep{lambourne2021brepnet} format by software. However, this process remains heavily reliant on manual input, making it time-consuming, particularly for intricate designs.

\begin{figure*}[t]
    \begin{center}
      \includegraphics[width=\linewidth]{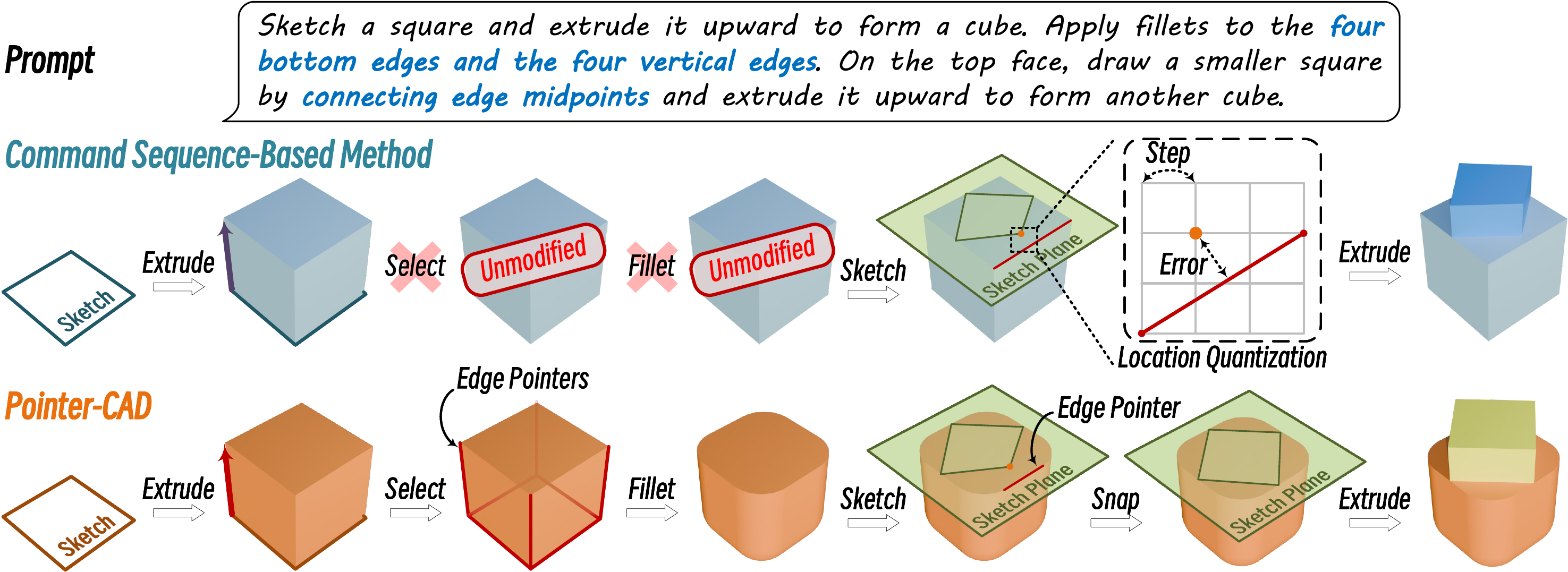}
    \end{center}
    \caption{\small 
        \textbf{Illustration of the strength of our proposed pointer-based command sequence compared to the previous command sequence-based CAD representation.} Command sequences suffer from the inability to refer to specific edges or faces, and discretization-induced quantization errors. In contrast, Pointer-CAD leverages edge pointers to directly refer to B-rep entities, enabling precise operations such as \textit{sketch} snapping, thereby reducing quantization errors and faithfully following complex text instructions.
    }
    \label{fig:teaser}
\end{figure*}

Recent efforts \citep{wu2025cmt,xu2024brepgen, alam2024gencad, xu2022skexgen} in CAD generation have explored parametric design synthesis with large generative models, aiming for fully autonomous CAD creation in an autoregressive manner.
Inspired by the reasoning capabilities of large language models (LLMs) \citep{achiam2023gpt,yang2025qwen3}, recent works \citep{xu2024cad, khan2024cad, you2024img2cad, alrashedy2024generating, wang2025cad, li2025cad} leverage LLMs or multimodal LLMs (MLLMs) to generate CAD models from natural language or other input modalities. These approaches can be broadly categorized into two lines: command sequence generation and code generation.
Code generation approaches~\cite{kolodiazhnyi2025cadrille,xie2025text} produce executable CAD scripts (e.g., in CadQuery~\cite{cadquery}), which can flexibly support a wide range of operations. While this flexibility is appealing, it comes at the cost of longer token sequences and higher inference time, as demonstrated in Table~\ref{tab:rep_statistics}.
In contrast, command sequence approaches~\cite{wu2021deepcad, khan2024text2cad, xu2024cad} encode CAD operations as sequences of tokens.
Their shorter length enables faster autoregressive generation and lower memory use, especially benefiting large-scale and interactive CAD generation tasks.
The main limitation of current command sequence methods, however, is the restricted set of supported editing operations. As shown in Figure \ref{fig:teaser}, operations such as \textit{chamfer} and \textit{fillet} refine existing geometry and require explicit selection of entities, which existing sequences handle poorly. Discretization of continuous variables further introduces quantization errors that can disrupt topological fidelity.

To address these limitations, inspired by Pointer Networks \citep{vinyals2015pointer}, we propose a pointer-based representation that explicitly references B-rep elements (e.g., edges and faces). This design mimics an engineer’s interaction with CAD software, enabling direct faces/edges selection and extending operations such as \textit{chamfer} and \textit{fillet}
, which are crucial in industrial CAD modeling. Moreover, by snapping predictions to referenced B-rep elements indicated by these pointers, our representation can mitigate coordinate errors from regression or quantization.
Building on the proposed pointer, we introduce a novel LLM-based text-to-CAD framework, \textbf{Pointer-CAD}.
Unlike prior approaches that generate full CAD models in a single step, Pointer-CAD adopts a multi-step strategy by decomposing the model into distinct steps: at each step, the B-rep from previous steps and the textual description condition the LLM to generate the parametric subsequent components.
Specifically, we extract geometric cues from B-rep faces and edges, construct a face-adjacency graph $\mathcal{G}$, and use  graph neural networks (GNNs) \citep{scarselli2008graph} to aggregate local features from neighboring elements.
Leveraging the reasoning capabilities of large language models, our framework outputs three complementary components, \textit{Label Tokens, Value Tokens}, and \textit{Pointer}, which can be directly translated into executable commands of CAD models.
When an operation requires geometric dependency on a previously generated structure, such as applying a \textit{chamfer} to an existing edge, the \textit{Pointer} is activated to select the most feature matching candidate face or edge.

To facilitate the performance evaluation of text-to-CAD generation, we design a CAD annotation pipeline by leveraging Qwen2.5-VL~\citep{bai2025qwen25vl} to generate high-level textual descriptions from multi-view CAD renderings. Building on the re-captioned OmniCAD dataset~\cite{xu2024cad} and further extending it with \textit{chamfer} and \textit{fillet} operations, we obtain a total of 575,559 models. For fair comparison with existing baselines, we adopt the DeepCAD~\cite{wu2021deepcad} split from this re-captioned dataset.
Our Pointer-CAD achieves strong performance on text-conditioned CAD generation, improving command sequence accuracy, and geometric reconstruction fidelity. Notably, the segment-level topological fidelity, quantified by the Segment Error (SegE), as well as the watertightness and enclosure quality of the solid, measured by the FluxEE metric, both show significant improvements compared with previous methods~\citep{khan2024text2cad,govindarajan2025cadmium}.

To conclude, our contributions can be summarized as follows: (1) A pointer-based command sequence representation enabling edge and face selection. This makes advanced operations like \textit{chamfer} and \textit{fillet} feasible for autoregressive methods and reduces quantization errors;
(2) We introduce Pointer-CAD, an LLM-based text-to-CAD framework built on the proposed representation. It employs a multi-step generation strategy, where each step is conditioned on both the textual description and the B-rep generated from previous steps; (3) Pointer-CAD outperforms existing baselines on text-conditioned generation, demonstrating superior reconstruction quality and topological consistency.

\section{Related Work}
\subsection{Boundary Representation}
Boundary Representation (B-rep) \citep{ansaldi1985geometric} uses a tree structure to organize vertices, edges, and faces in a hierarchical way.
Several methods generate B-reps by progressively constructing these hierarchical structures \citep{nash2020polygen, guo2022complexgen, jayaraman2022solidgen}.
Additionally, some recent approaches~\citep{xu2024brepgen, liu2025hola} leverage latent spaces to encode the complex topology of B-reps. Based on these B-rep latent representation, CMT~\citep{wu2025cmt} takes an effort to utilize a continuous autoregressive manner for B-rep generation. Although B-reps provide a direct representation of 3D models, the intricate relationships among elements make them challenging to generate.

\subsection{Constructive Solid Geometry}
Constructive Solid Geometry (CSG) \citep{foley1996computer} represents objects by combining primitive shapes through Boolean operations.
Due to the non-uniqueness of CSG representations, researchers often employ unsupervised training methods \citep{sharma2018csgnet, du2018inversecsg, kania2020ucsg}.
Recent works propose CSG-like representations \citep{ren2021csg} and learnable primitives \citep{yu2023d, yu2022capri} to improve generation quality.
However, CSG methods struggle to represent curved surfaces such as rounded corners, limiting their capacity for complex geometries.

\subsection{Command Sequence Representation}
With the emergence of large-scale CAD datasets \citep{koch2019abc, willis2021fusion}, deep models for CAD generation have advanced significantly. DeepCAD \citep{wu2021deepcad} encodes design parameters as command sequences, SkexGen \citep{xu2022skexgen} integrates primitive hierarchies for autoregressive generation, and TransCAD \citep{dupont2024transcad} uses hierarchical structures to enhance geometric reasoning. Token-based diffusion models further enable sequence generation \citep{ma2024draw, zhang2025diffusion, yu2026gencad}. Recent studies explore large language models (LLMs) to generate CAD sequences from point clouds \citep{khan2024cad}, images \citep{chen2025img2cad}, and text \citep{khan2024text2cad}. CAD-MLLM \citep{xu2024cad} proposes a multi-modal LLM framework that integrates these three modalities, and CAD-GPT~\citep{wang2025cad} integrates images and text. FlexCAD \citep{zhang2024flexcad} enables controllable generation, and CADFusion \citep{wang2024cadfusion} leverages visual feedback to improve sequence quality. Despite these advances, command sequences generally lack explicit topological information, which remains a key challenge for autoregressive generation. A recent work~\citep{fan2025parametric} attempts to enable entity selection by labeling faces based on each operation and edges as intersections of faces. However, edges derived from face intersections may not be unique, leading to ambiguity in the selection function. A more robust solution for entity selection remains an open problem.

\subsection{Code Representation}
With the rise of open-source pretrained code models \citep{huang2025opencoder, zheng2023codegeex}, several approaches \citep{govindarajan2025cadmium, guan2025cad, li2025cad, niu2025creft, yuan2024openecad, rukhovich2025cad, kolodiazhnyi2025cadrille} represent the modeling process directly as plain-text code representation to simplify fine-tuning for LLMs. Among these, CADmium \citep{govindarajan2025cadmium} uses \textit{JSON} code, whereas \citep{guan2025cad, li2025cad, niu2025creft, yuan2024openecad, rukhovich2025cad, kolodiazhnyi2025cadrille} employ \textit{Python} code via CadQuery \citep{cadquery} or the FreeCAD \citep{freecad} API. Some works further enhance the geometric reasoning of LLMs using Chain-of-Thought (CoT) prompting \citep{niu2025creft}. However, these representations are generally not optimized for compression, so the token sequences representing a CAD model can remain relatively long as presented in Figure~\ref{tab:rep_statistics}, which would potentially reduce both training and inference efficiency.

\begin{table}
  \small
  \caption{
    \textbf{Comparison of Different Representations.}
    Length indicates the average number of tokens required to express a CAD model using each representation, and Time denotes the average generation time for producing a model.
  }
  \label{tab:rep_statistics}
  \centering
  \begin{adjustbox}{max width=\columnwidth}
\begin{tabular}{ccccc}
\toprule
Method & Base Model & Representation Type & Length (token) & Time (s) \\
\midrule
Pointer-CAD & Qwen2.5-0.5B & Command Sequence & 110.72  & 2.13  \\
Pointer-CAD & Qwen2.5-1.5B & Command Sequence & 110.72  & 2.71  \\
Text2CAD & Transformer-363M & Command Sequence & 43.97  & 1.61  \\
Text-to-CadQuery & Qwen2.5-0.5B & CadQuery Code & 424.75  & 6.14  \\
Text-to-CadQuery & Qwen2.5-1.5B & CadQuery Code & 424.75  & 6.61  \\
cadrille & Qwen2-2B & CadQuery Code & 424.75  & 4.72  \\
\bottomrule
\end{tabular}%

  \end{adjustbox}
\end{table}

\section{Pointer-Based Command Sequences}
\label{sec:representation}
Contemporary CAD software allows direct selection of operation targets on rendered geometry (e.g., clicking an edge for chamfering). In contrast, prior works \citep{wu2021deepcad, khan2024text2cad, xu2024cad, govindarajan2025cadmium} represent operations purely as numerical sequences, ignoring previous geometric context from previous steps.
As shown in Figure \ref{fig:teaser}, this leads to two key issues: (i) operations like \textit{chamfer} and \textit{fillet} remain unsupported because they require explicit geometric entity references;
(ii) quantization errors inherent in LLM-based sequential generation cause newly drawn curves to fail to snap to existing edges, and sketch planes to misalign with target faces, introducing small errors that hinder precise geometric connectivity or alignment during sequential generation.
Motivated by Pointer Networks \citep{vinyals2015pointer}, we propose a novel pointer-based command sequence representation that explicitly integrates B-rep geometry into sequential modeling.

In our representation, each token belongs to one of three types: \textit{Label Token}, \textit{Value Token}, or \textit{Pointer}.
The \textit{Label Token} carries explicit semantic information, indicating the type of an operation or a structural boundary in the sequence, as detailed in Table \ref{tab:token}.
The \textit{Value Token} provides numerical data, such as coordinates or degrees. Notably, the continuous parameters are quantized into $2^q$ levels and expressed as $q$-bit integers.
The \textit{Pointer} is used to reference a face or an edge from the B-rep.
Different operations are then defined by specific combinations and sequential order of these tokens. We decompose the entire CAD model construction process into a sequence of steps, each consisting of one of three fundamental operations: a sketch-extrude combination, a chamfer, or a fillet. And a CAD model is then represented by an ordered sequence of these operations.
\begin{table}[tb]
    \small
    \caption{\small
        \textbf{Label Token Definitions.}
        This table provides a comprehensive list of all \textit{Label Tokens} used in our command sequence representation, along with their semantic descriptions.
    }
    \label{tab:token}
    \begin{adjustbox}{max width=0.95\linewidth}
\begin{tabular}{ll|ll}
\toprule
Command & Description & Command & Description \\
\midrule
\commandnotation{ss} & Start of sketch & \commandnotation{sx} & Start of curve \\
\commandnotation{se} & Start of extrusion & \commandnotation{or} & Orientation token  \\
\commandnotation{sc} & Start of chamfer & \commandnotation{dr} & Direction token\\
\commandnotation{sf} & Start of fillet & \commandnotation{bo} & Boolean token \\
\commandnotation{sp} & Start of profile & \commandnotation{em} & End of model  \\
\commandnotation{sl} & Start of loop & \commandnotation{es} & End of step  \\
\bottomrule
\end{tabular}%

    \end{adjustbox}
\end{table}

\textbf{Sketch-extrude combination step.}
Following prior works ~\citep{wu2021deepcad,govindarajan2025cadmium,khan2024text2cad}, we define a 2D sketch hierarchically: a sketch consists of faces, each bounded by one or more loops. A loop is formed by a sequence of primitive curves (lines or arcs) or a single circle, with consecutive curves sharing endpoints. The primitives are parameterized as:
(i) $Line : (x, y)$, where $(x, y)$ defines the start point of a line; (ii) $Arc: (x, y, \alpha, o)$ which defines an arc with the start point $(x, y)$ and sweep angle $\alpha$, and $o$ refers to the orientation flag (denoted as \commandnotation{or}); (iii) $Circle: (x, y, r)$, where $(x, y)$ is the center of an circle with a radius $r$. 

For sketch plane selection, we replace the conventional six-parameter representation (three Euler angles and three translation) with a pointer mechanism that directly selects a target face from the B-rep representation to serve as the sketch plane. A local 2D coordinate system is then established on this plane, providing a consistent reference frame for all subsequent sketch operations (see Supplementary for construction details).
This pointer-based approach reformulates plane selection from a 3D rotation regression problem into a discrete selection over a finite set of candidate faces, reducing the search space and mitigating misalignment caused by inaccurate regression or quantization errors.

With the sketch plane fixed, \textit{extrude} is simplified as $E:(e_{p}, e_{n}, b)$, where $e_{p}$ and $e_{n}$ denote extrusion distances along the positive and negative normal directions, and $b$ (denoted as \commandnotation{bo}) specifies the boolean type (e.g., \textit{New}, \textit{Join}, \textit{Cut}, \textit{Intersect}).

\textbf{\textit{Chamfer} or \textit{fillet} operations step.}
Mirroring the workflow in modern CAD software, both operations first require the selection of one or more target edges, then assign a single numerical parameter.
We represent \textit{chamfer} as $C:(\mathbf{p}, c)$ and \textit{fillet} as $F:(\mathbf{p}, f)$, where $\mathbf{p} = \{p_1, p_2, \dots, p_n\}$ is a set of pointers, with each pointer $p_i$ identifying a target edge from the B-rep.
The parameters $c$ and $f$ denote the chamfer distance and fillet radius, which are applied uniformly across all selected edges.
\section{Method}

\begin{figure*}[thb]
    \begin{center}
        \includegraphics[width=0.85\linewidth]{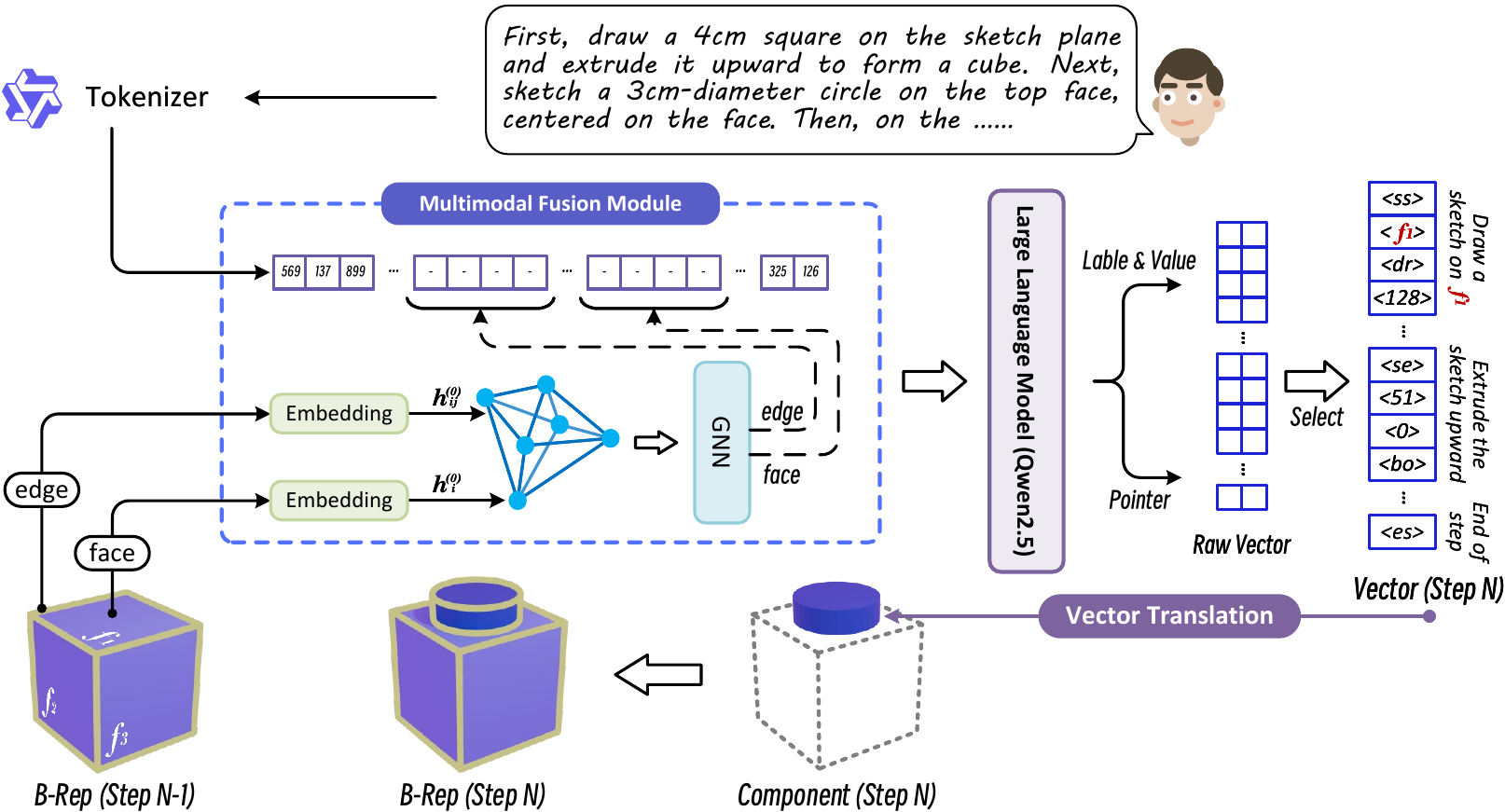}
    \end{center}
    \caption{\small
\textbf{Pointer-CAD Pipeline.} At each generation step, the full user prompt is tokenized, while the B-rep is updated with all geometry generated so far. A multimodal fusion module combines the textual prompt with the evolving B-rep, which is further encoded via a graph neural network over its faces and edges. The fused representation is then processed by a large language model to predict the vector for the current step, which is subsequently translated into geometry to update the B-rep.
    }
    \label{fig:pipeline}
\end{figure*}

Building on the proposed pointer-based command sequence, we introduce \MethodName{}, a framework that transforms text descriptions into 3D CAD models. 
In addition, we introduce an annotation pipeline and construct a new dataset to fully unleash the potential of \MethodName{}.
This section details the overall architecture, training objectives, and the annotation pipeline.

\subsection{Overall Architecture}
As illustrated in Figure \ref{fig:pipeline}, unlike previous command sequence approaches that treat generation as a whole sequence, we separate the process into multiple steps that are predicted sequentially in an autoregressive manner.
Each prediction conditions on the text description and the B-rep geometry accumulated so far, ensuring global consistency and faithful design semantics.
\MethodName{} comprises three key components: a Multimodal Fusion Module that integrates text and B-rep geometry, an LLM for sequence generation, and a Vector Translation Module that converts command sequences into B-rep representations following the construction process described in Section \ref{sec:representation}.

\subsubsection{Multimodal Fusion Module}
The Multimodal Fusion Module integrates tokenized text and B-rep geometry to provide structured representations for subsequent processing. The text is tokenized once and reused across all steps, while the B-rep is incrementally updated after each operation. At the first step, the B-rep is empty, so the model conditions only on the text.

\textbf{B-rep Encoder.}
We represent the B-rep as an undirected face-adjacency graph $\mathcal{G}(\mathit{V}, \mathit{E})$, where nodes denote faces and edges denote shared boundaries. Following \citep{jayaraman2021uv, yin2025rlcad}, we build the initial graph $\mathcal{G}$ by sampling geometric cues from the parametric domains of B-rep faces and edges.
Each face $\mathcal{S}(u,v)$ is uniformly sampled on a $32{\times}32$ grid in the $(u,v)$ domain, with 3D coordinates, surface normals, Gaussian curvature, and visibility indicator concatenated as features.
Similarly, each edge $\mathcal{C}(t)$ is uniformly sampled with 32 points, extracting the 3D coordinates, tangent and its reverse vector, and first-order derivative.
Point-wise features are aggregated via average pooling and projected to a 128-d embedding, yielding node features $h_i^{(0)}$, and the edge feature $h_{ij}^{(0)}$, where $i,j$ are the indices of the faces for the initial graph $\mathcal{G}$. 
Further details are in the Supplementary.

\textbf{Graph Processing.}
After obtaining the initial features, we take into account the structural properties of the B-rep and apply a $K$-layer Graph Neural Network (GNN)\citep{scarselli2008graph} to propagate information. 
Node features are updated by aggregating messages from neighboring edges, while edge features are updated using both the features of their incident nodes and a multi-head attention (MHA) mechanism \citep{vaswani2017attention} over all node features to capture global dependencies.

At the $k$-th layer, the updates are formulated as:

{\footnotesize
\begin{align*}
h_{i}^{(k)} &= \phi^{(k)} \Bigl( (1+\epsilon^{(k)})\, h_{i}^{(k-1)} 
               + \sum_{j \in \mathcal{N}(i)} f_{\Theta}(h_{ij}^{(k-1)}) \odot h_{j}^{(k-1)} \Bigr), \\[0.3ex]
h_{ij}^{(k)} &= h_{ij}^{(k-1)} + \mathrm{MHA}\Bigl( Q = h_{ij}^{(k-1)},\; 
                K,V = \{ h_{l}^{(k-1)} \mid l \in \mathcal{V} \} \Bigr) \\
             &+ \psi^{(k)} \Bigl( (1+\gamma^{(k)})\, h_{ij}^{(k-1)} 
               + f_{\Xi}(h_{i}^{(k-1)} + h_{j}^{(k-1)}) \Bigr),
\end{align*}}
where $\phi^{(k)}, \psi^{(k)}$ are MLPs, $\epsilon^{(k)}, \gamma^{(k)}$ are learnable scalars, and $f_{\Theta}, f_{\Xi}$ project features between edge and node spaces.
The resulting node and edge embeddings, $h_i^{(k)}$ and $h_{ij}^{(k)}$, are serialized into the LLM input via structured prompting: edge embeddings are wrapped as \textit{\commandnotation{brep\_edge\_start} edge embedding \commandnotation{brep\_edge\_end}} and face embeddings as \textit{\commandnotation{brep\_face\_start} face embedding \commandnotation{brep\_face\_end}}, enabling the LLM to distinguish B-rep components.

\subsubsection{Supervised Finetuning of Large Language Models}
LLMs exhibit strong reasoning over structured inputs. In Pointer-CAD, we adopt Qwen2.5 \citep{team2024qwen2} as the backbone and apply Low-Rank Adaptation (LoRA) \citep{hu2022lora} to reduce trainable parameters. To align with our representation, we append two separate fully connected layers to the final hidden state: one predicts the \textit{Label Token} and \textit{Value Token}, while the other predicts the \textit{Pointer}.
Outputs are then translated into executable command sequences following the rules detailed in the Supplementary. 

\textbf{Pointer-based Referencing.}
In the pointer-enabled setting, the LLM predicts a \textit{Pointer} to select the target face or edge from candidate sets.
We denote the all faces (including the three base planes: Right, Front, and Top) and edges as $\mathcal{S}_f$ and $\mathcal{S}_e$.
Since geometric relations (e.g., coplanar faces, collinear edges) may yield multiple valid targets, the ground-truth pointer is defined as a subset.
Precise definitions of these geometric special cases are provided in the Supplementary.
Formally, for the $m$-th predicted face pointer, we define the ground-truth set as $\mathcal{P}_m \subseteq \mathcal{S}_f$, and $\mathcal{N}_m = \mathcal{S}_f \setminus \mathcal{P}_m$ is negative. Similarly,  $\mathcal{P}_n \subseteq \mathcal{S}_e$ and $\mathcal{N}_n = \mathcal{S}_e \setminus \mathcal{P}_n$ for the $n$-th predicted edge pointer.
Each candidate uses its initial feature: $h_i^{(0)}$ for the $i$-th face in $\mathcal{S}_f$, and $h_{ij}^{(0)}$ for the edge shared by the $i$-th and $j$-th faces in $\mathcal{S}_e$, with three base planes encoded as distinct learnable 128-d embeddings, aligning with features in both $\mathcal{S}_f$ and $\mathcal{S}_e$.
To predict a face or edge pointer, the LLM outputs a 128-d vector, which is matched to the candidate geometric element with highest cosine similarity. 

\begin{figure*}[t]
    \begin{center}
       \includegraphics[width=0.8\linewidth]{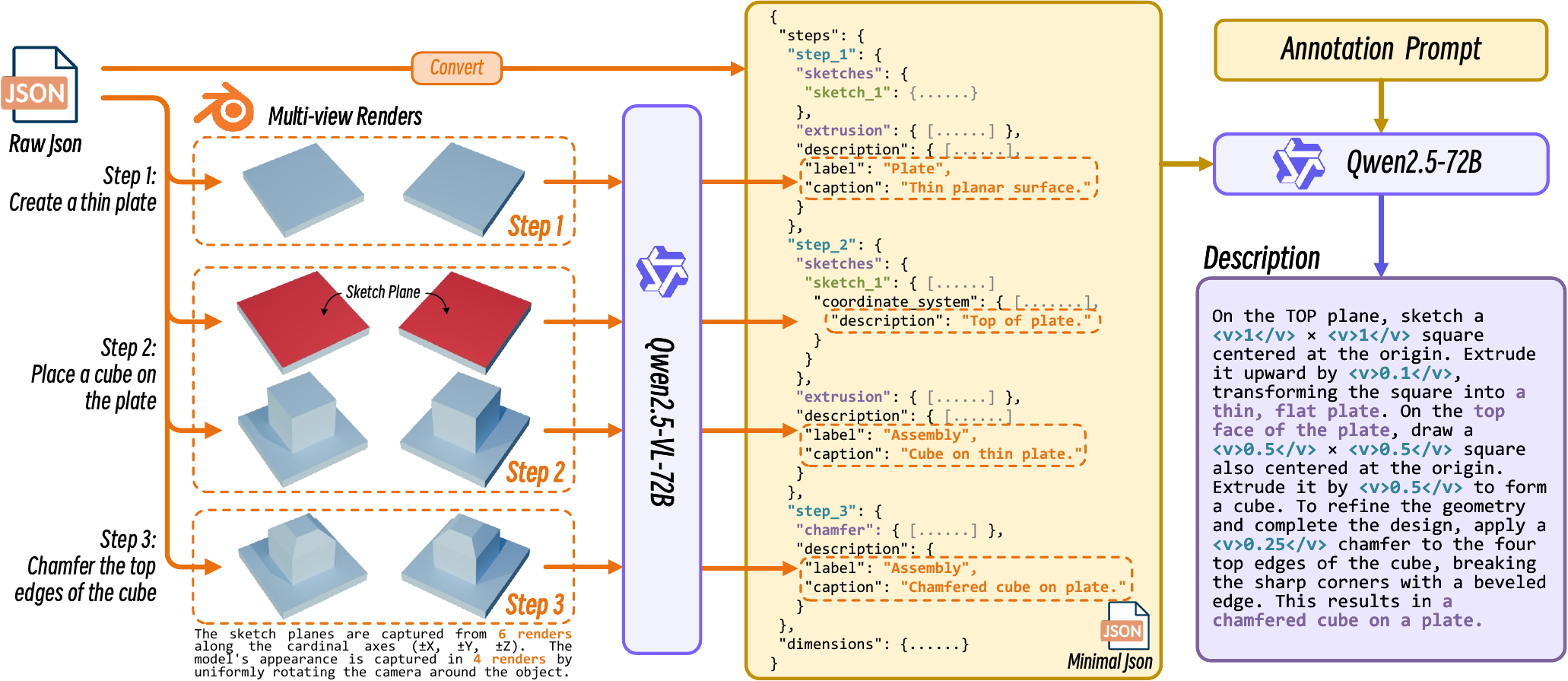}
    \end{center}
    \caption{\small 
        \textbf{Dataset construction pipeline.}
        Raw JSONs are converted into a minimal format containing only annotation-relevant elements. Sketch planes and models are rendered, and Qwen2.5-VL generates textual descriptions for integration into the JSON. Finally, Qwen2.5 produces step-by-step instructions, with dimension parameters wrapped in special tags for future data augmentation.
    }
    \label{fig:annotation}
\end{figure*}

\subsection{Training Objective}
\label{sec:training_objective}

Based on the structure of the command sequence, our training objective is to jointly predict the correct token value and referenced pointer representation.

\textbf{Label and Value Token Prediction.}
The prediction of both \textit{Label Tokens} and \textit{Value Tokens} is formulated as a classification task.
Given the constrained output space, we employ a cross-entropy loss with label smoothing, defined as:
{\small
\begin{equation*}
\mathcal{L}_v = -\sum_{i=1}^N \left[ (1 - \alpha) \cdot \delta_{i,y} + \frac{\alpha}{N - 1} \cdot (1 - \delta_{i,y}) \right] \log p_i ,
\end{equation*}}
where $\delta_{i,y}$ is the Kronecker delta (1 if $i = y$, 0 otherwise), $y$ is the correct class, $N$ is the number of classes, $\alpha$ is the label smoothing factor, and $p_i$ is the predicted probability of class $i$, obtained via softmax over the model logits. 

\textbf{Pointer Prediction.}
Pointer prediction is cast as a regression task.
Since multiple valid pointers may exist simultaneously, we adopt a contrastive-style loss:
{\small
\begin{equation*}
\begin{split}
\mathcal{L}_p = {}& -\frac{1}{|\mathcal{P}| + |\mathcal{N}|} \biggl[ 
\sum_{j \in \mathcal{P}} \log \left( \sigma\!\left( \frac{ \cos(p, c_j) }{ \tau } \right) \right) \\[0.8ex]
& {}+ \sum_{j \in \mathcal{N}} \log \left( 1 - \sigma\!\left( \frac{ \cos(p, c_j) }{ \tau } \right) \right) 
\biggr],
\end{split}
\end{equation*}}
where $\mathcal{P}$ and $\mathcal{N}$ denote the sets of valid and invalid candidates, $p$ is the predicted pointer embedding, $c_j$ is the embedding of candidate $j$, $\sigma$ is the sigmoid function, and $\tau$ is a learnable temperature. 

\textbf{Overall Objective.}
The overall loss is a weighted sum of the two objectives:
\begin{equation*}
\mathcal{L} = \lambda_v \cdot \mathcal{L}_v + \lambda_p \cdot \mathcal{L}_p ,
\end{equation*}
where $\lambda_v$ and $\lambda_p$ are hyperparameters controlling the relative contributions of these two components.

\begin{table*}
  \small
  \caption{
    \textbf{Quantitative comparison on different datasets.}
    (a) Recap-DeepCAD dataset: Pointer-CAD (0.5B/1.5B) achieves the highest operation F1 scores and lowest CD errors, outperforming other baselines and larger LLM-based method CADmium-7B.
    (b) Recap-OmniCAD$^+$ dataset: Pointer-CAD uniquely supports \textit{chamfer} and \textit{fillet} operations with high accuracy, while other methods fail, and further demonstrates superior geometric fidelity and topology quality. 
  }
  \label{tab:performance}
  \centering
  \begin{adjustbox}{max width=\linewidth}
\begin{tabular}{cccccccc|ccccccc}
\toprule
      & \multicolumn{7}{c|}{(a) Recap-DeepCAD}                & \multicolumn{7}{c}{(b) Recap-OmniCAD\textsuperscript{+}} \\
\cmidrule{2-15}\multirow{2}[4]{*}{Model} & \multirow{2}[4]{*}{DeepCAD} & \multirow{2}[4]{*}{Text2CAD} & \multicolumn{3}{c}{CADmium} & \multicolumn{2}{c|}{Pointer-CAD} & \multirow{2}[4]{*}{DeepCAD} & \multirow{2}[4]{*}{Text2CAD} & \multicolumn{3}{c}{CADmium} & \multicolumn{2}{c}{Pointer-CAD} \\
\cmidrule{4-8}\cmidrule{11-15}      &       &       & 1.5B  & 3B    & 7B    & 0.5B  & 1.5B  &       &       & 1.5B  & 3B    & 7B    & 0.5B  & 1.5B \\
\midrule
Line F1 \textuparrow & 80.14  & 88.12  & 85.47  & 82.25  & 85.13  & 97.70  & \textbf{98.73 } & 83.10  & 88.61  & 84.91  & 82.84  & 84.94  & 94.37  & \textbf{95.79 } \\
Arc F1 \textuparrow & 31.41  & 45.19  & 19.35  & 20.44  & 25.68  & 85.70  & \textbf{95.14 } & 29.69  & 42.00  & 20.89  & 24.68  & 27.25  & 67.62  & \textbf{74.98 } \\
Circle F1 \textuparrow & 79.04  & 87.03  & 75.64  & 72.66  & 74.94  & 98.27  & \textbf{98.66 } & 66.53  & 78.54  & 67.52  & 65.60  & 65.26  & 95.61  & \textbf{96.03 } \\
Extrusion F1 \textuparrow & 92.34  & 98.53  & 92.50  & 88.50  & 90.75  & \textbf{99.67 } & 99.61  & 92.45  & 94.27  & 93.55  & 89.73  & 91.56  & \textbf{99.22 } & 99.20  \\
Chamfer F1 \textuparrow & -     & -     & -     & -     & -     & -     & -     & -     & -     & -     & -     & -     & 89.74  & \textbf{94.32 } \\
Fillet F1 \textuparrow & -     & -     & -     & -     & -     & -     & -     & -     & -     & -     & -     & -     & 82.54  & \textbf{89.85 } \\
CD mean \textdownarrow & 37.47  & 17.48  & 11.51  & 12.22  & 10.53  & 3.81  & \textbf{2.58 } & 27.48  & 12.56  & 13.24  & 13.29  & 11.60  & 5.49  & \textbf{2.86 } \\
CD median \textdownarrow & 12.56  & 3.38  & 0.57  & 0.47  & 0.44  & 0.54  & \textbf{0.30 } & 11.37  & 3.67  & 1.02  & 0.98  & 0.82  & 0.53  & \textbf{0.34 } \\
SegE \textdownarrow & 0.53  & 0.44  & 0.47  & 0.64  & 1.21  & 0.13  & \textbf{0.11 } & 0.89  & 0.51  & 0.81  & 0.85  & 1.39  & \textbf{0.15 } & 0.17  \\
FluxEE \textdownarrow & 25.85  & 17.75  & 38.63  & 29.73  & 32.22  & \textbf{2.14 } & 2.97  & 42.59  & 26.36  & 35.16  & 32.03  & 36.30  & 3.51  & \textbf{3.44 } \\
\bottomrule
\end{tabular}%
  \end{adjustbox}
\end{table*}

\subsection{Annotation Pipeline}

As shown in Figure~\ref{fig:annotation}, we render four multi-view images per model using Blender and use Qwen2.5-VL \citep{bai2025qwen25vl} to generate a one-word label and single-sentence caption for global shape understanding.
For each sketch plane, six views are rendered with the plane highlighted in red, and Qwen2.5-VL generates a macro-level spatial description. 
These annotations provide a more comprehensive understanding of both the model geometry and the sketch plane location.
We convert raw JSON files into a concise, human-readable JSON format, enhanced with textual descriptions for better interpretability.
Unlike Text2CAD \citep{khan2024text2cad}, which removes units and normalizes geometry, we preserve actual parameters to reflect the true construction process.
We also employ Qwen2.5 \citep{Yang2024Qwen25} to generate modeling instructions, wrapping all dimension values in \commandnotation{v} tags.
Without normalization, parameters no longer correspond to a canonical reference shape as in Text2CAD, shifting the normalization burden to downstream models and making the task more challenging.
Building on OmniCAD \citep{xu2024cad}, we adopt its stage-wise augmentation by splitting full models into intermediate sub-models and annotating each via our pipeline, forming Recap-OmniCAD. 
Since \textit{chamfer} and \textit{fillet} were absent in OmniCAD, we reintegrate them and extend the dataset to create OmniCAD\textsuperscript{+} with new captions.

\section{Experiments}

\subsection{Experimental Setup}

\textbf{Datasets.} 
To validate Pointer-CAD and ensure fair comparison, we additionally annotate a subset of the DeepCAD dataset, denoted as Recap-DeepCAD, containing 176,439 CAD models.
To evaluate support for \textit{chamfer} and \textit{fillet} operations, we train on Recap-OmniCAD\textsuperscript{+} dataset, with a total number of 575,559. 
Additional statistics and annotation prompts are provided in the Supplementary.

\textbf{Implementation Details.}
Unless otherwise specified, we use Qwen2.5-0.5B \citep{Yang2024Qwen25} as the backbone LLM for \MethodName{}. 
The model is trained for 10 epochs and more training details are included in the Supplementary.

\textbf{Metrics.} 
Following Text2CAD \citep{khan2024text2cad}, CAD-MLLM \citep{xu2024cad}, and CADmium \citep{govindarajan2025cadmium}, we report the F1 score, Chamfer Distance (CD), Segment Error (SegE), and Flux Enclosure Error (FluxEE).
The F1 score measures command accuracy, CD evaluates geometric fidelity, SegE reflects topological correctness, and FluxEE quantifies deviation from watertight solids. 
All experiments are conducted in a normalized space of $[-0.5, 0.5]^3$ for consistent spatial alignment.
CD is computed with 8,192 sampled points, and both CD and FluxEE are scaled by $10^3$ for readability. Additional metrics are reported in the Supplementary.  

\begin{figure}[t]
    \begin{center}
        \includegraphics[width=\linewidth]{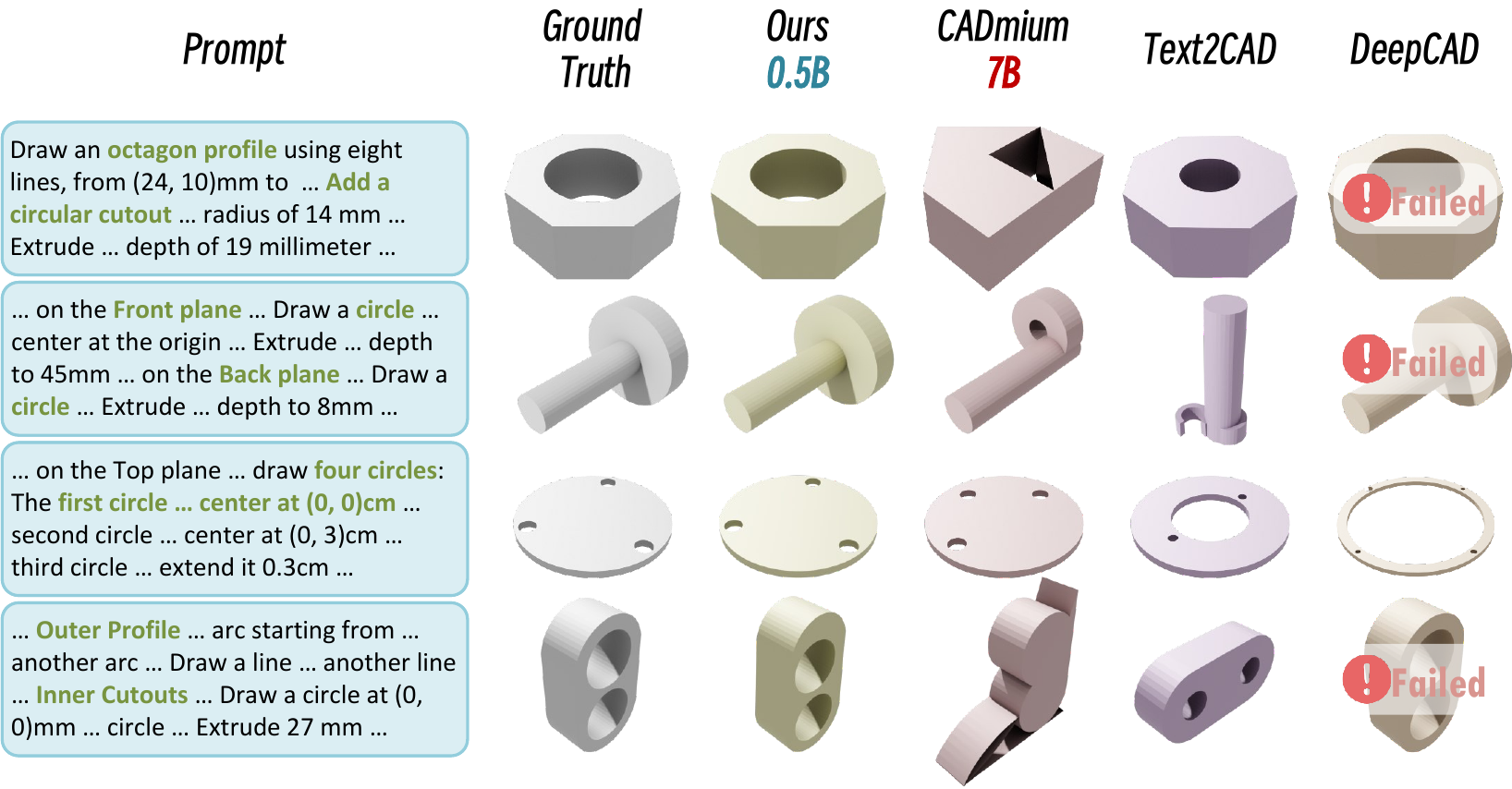}
    \end{center}
    \caption{\small 
    \textbf{Qualitative performance comparison on Recap-DeepCAD dataset.}
    Our method consistently produces accurate and faithful geometry aligned with the ground truth, while competing methods often miss details or collapse entirely. Notably, Pointer-CAD achieves superior results among LLM-based methods despite a significantly smaller size than CADmium.
    }
    \label{fig:performance}
\end{figure}

\subsection{Comparison on Text-to-CAD Generation}
We involve two open-source text-to-CAD baselines for comparison: Text2CAD~\citep{khan2024text2cad} and CADmium~\citep{govindarajan2025cadmium}. 
In addition, we adapt DeepCAD~\citep{wu2021deepcad} for text-conditioned generation by reusing its pretrained latent-space decoder, which was trained on the DeepCAD dataset, and training a new encoder to map text inputs into the corresponding latent vectors. To ensure fair comparison, all the baseline methods and \MethodName{} are trained on Recap-DeepCAD dataset, which excludes operations unsupported by some methods.
As shown in Table \ref{tab:performance}(a), \MethodName{}-1.5B achieves the best sketch operation F1, CD and SegE, while \MethodName{}-0.5B attains the best performance on the remaining metrics.
Notably, our method achieves a substantially smaller SegE than all baselines, demonstrating that the proposed pointer mechanism effectively mitigates discontinuities and connectivity problems caused by even tiny quantization errors, which can separate newly drawn parts from existing geometry.
Moreover, our method attains superior overall performance with a 0.5B model size compared to the 7B-LLM-based CADmium. 
Additional quantitative results, evaluated across more datasets and metrics, are provided in the Supplementary for further analysis.

To further assess our model's capability on \textit{chamfer} and \textit{fillet} operations, we train our model on the \Dataset{} dataset, which includes these two operation types. Baselines, lacking support for these operations, are trained on Recap-OmniCAD instead. Table~\ref{tab:performance}(b) shows our approach faithfully reconstructs these operations, achieving better geometric accuracy, and superior topology quality.

Additionally, we compare with state-of-the-art general-purpose LLMs (Claude~\cite{anthropic2025claudeopus4}, Gemini~\cite{google2025gemini25pro}, GPT~\cite{openai2025gpt52}, Qwen~\cite{yang2025qwen3}) to highlight the necessity of specialized CAD architectures. Prompting them to generate CadQuery code on 2K randomly selected subsets from both datasets, Table~\ref{tab:llm-inference} details that general LLMs struggle to produce executable, geometry-consistent codes. In contrast, \MethodName{} substantially outperforms them across all metrics.

\begin{figure}[thb]
    \begin{center}
        \includegraphics[width=\linewidth]{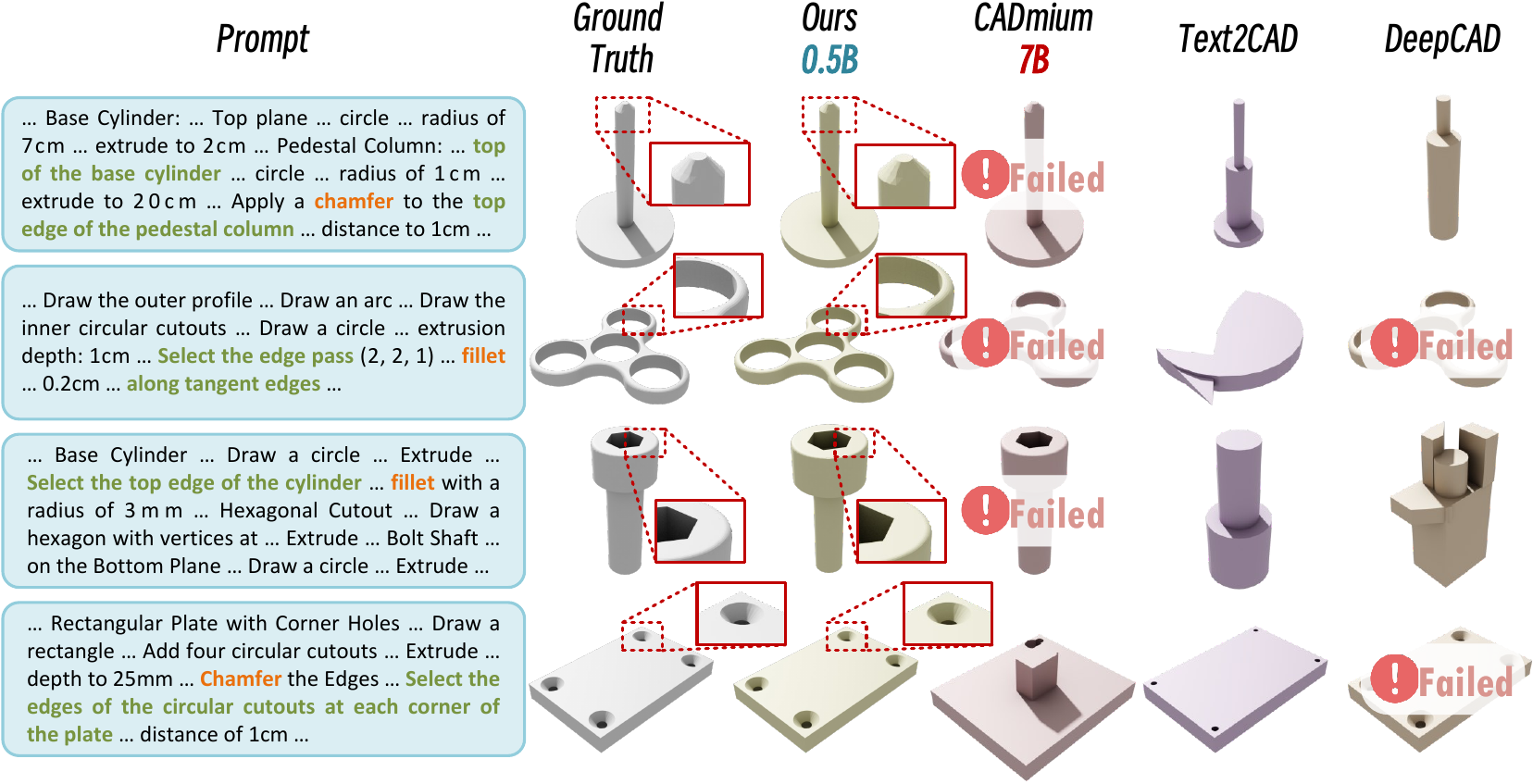}
    \end{center}
    \caption{\small
    \textbf{Qualitative performance comparison on \Dataset{} dataset.}
    Our method accurately recovers detailed structures that closely match the ground truth for complex CAD models involving \textit{chamfer} or \textit{fillet} operations. Conversely, competing methods often miss fine-grained features or fail entirely.
    }
    \label{fig:performance2}
\end{figure}

Qualitative comparisons validate these quantitative findings. As observed in Figure~\ref{fig:performance}, baseline methods frequently produce defective CAD models, exhibiting issues such as overly thin surfaces or incorrect spatial arrangement of internal structures. Furthermore, as illustrated in Figure~\ref{fig:performance2}, existing methods often fail to execute \textit{chamfer} and \textit{fillet} operations correctly, producing invalid results. In contrast, our method explicitly incorporates geometric information from B-rep into the modeling process, leading to significantly improved structural accuracy. Overall, our pointer-based representation and training strategy show strong compatibility with autoregressive models.

\begin{table}
  \setlength{\tabcolsep}{2pt}
  \renewcommand{\arraystretch}{0.85}
  \caption{\small 
    \textbf{Comparison with existing LLMs.}
    LLMs struggle to produce executable, geometry-consistent CadQuery codes.
  }
  \label{tab:llm-inference}
  \centering
  \begin{adjustbox}{max width=\linewidth}
\begin{tabular}{ccccccc|cccccc}
\toprule
  & \multicolumn{6}{c|}{(a) Recap-DeepCAD-2K} & \multicolumn{6}{c}{(b) Recap-OmniCAD\textsuperscript{+}-2K} \\
\cmidrule{2-13}Model & Claude & Gemini & GPT & Qwen3 & \multicolumn{2}{c|}{Pointer-CAD} & Claude & Gemini & GPT & Qwen3 & \multicolumn{2}{c}{Pointer-CAD} \\
\textit{version} & Opus 4 & 2.5 Pro & 5.2 & 235B-A22B & 0.5B & 1.5B & Opus 4 & 2.5 Pro & 5.2 & 235B-A22B & 0.5B & 1.5B \\
\midrule
IR \textdownarrow & 29.75  & 24.95  & 23.90  & 35.80  & 14.79  & \textbf{8.67 } & 41.75  & 39.25  & 33.55  & 49.70  & 26.02  & \textbf{19.11 } \\
CD mean \textdownarrow & 31.38  & 15.04  & 35.13  & 28.85  & 3.98  & \textbf{2.65 } & 31.03  & 14.94  & 35.08  & 29.54  & 5.43  & \textbf{2.92 } \\
CD median \textdownarrow & 6.31  & 0.58  & 9.69  & 1.80  & 0.54  & \textbf{0.28 } & 8.61  & 0.82  & 10.74  & 2.33  & 0.53  & \textbf{0.35 } \\
\bottomrule
\end{tabular}%

  \end{adjustbox}
\end{table}

\subsection{Ablation on the GNN component}
To verify the efficiency of the GNN component, we conduct a comparison in Table \ref{tab:ablation} across four settings: 1) Our method with the GNN replaced by a 3-layer MLP; 2) Our full GNN method;  3) The original Text2CAD baseline; 4) Text2CAD augmented with our GNN.
To integrate our GNN into Text2CAD, we adopt a multi-step generation strategy and encode the current B-rep with a GNN at the beginning of each step.
The GNN-based variant significantly outperforms the MLP baseline, suggesting that GNNs help the model better capture complex geometric structures, particularly for arc edges.

\begin{table}
    \setlength{\tabcolsep}{3pt}
    \renewcommand{\arraystretch}{0.85}
    \caption{\small
      \textbf{GNN ablation on the Recap-DeepCAD dataset.}
      The GNN yields consistent gains, particularly on arc structures.
    }
    \label{tab:ablation}
    \centering
    \begin{adjustbox}{max width=\linewidth}
\begin{tabular}{ccccccccc}
\toprule
\multicolumn{2}{c}{Model} & IR \textdownarrow & Line F1 \textuparrow & Arc F1 \textuparrow & Circle F1 \textuparrow & Extrusion F1 \textuparrow & CD mean \textdownarrow & CD median \textdownarrow \\
\midrule
Pointer-CAD & \textit{w/o} GNN & 22.73  & 94.64  & 67.14  & 96.32  & 99.51  & 5.13  & 0.72  \\
0.5B & \textit{w/} GNN & \textbf{15.02} & \textbf{97.70 } & \textbf{85.70 } & \textbf{98.27 } & \textbf{99.67 } & \textbf{3.81 } & \textbf{0.54 } \\
\midrule
\multirow{2}[2]{*}{Text2CAD} & \textit{w/o} GNN & 30.16  & 88.12  & 45.19  & 87.03  & 98.53  & 17.48  & 3.38  \\
  & \textit{w/} GNN & \textbf{27.17 } & \textbf{90.28 } & \textbf{51.85 } & \textbf{89.94 } & \textbf{99.08 } & \textbf{14.33 } & \textbf{2.11 } \\
\bottomrule
\end{tabular}%

    \end{adjustbox}
\end{table}

\subsection{Visualization of Complex Cases}
\begin{figure}[t]
  \centering
  \includegraphics[width=\linewidth]{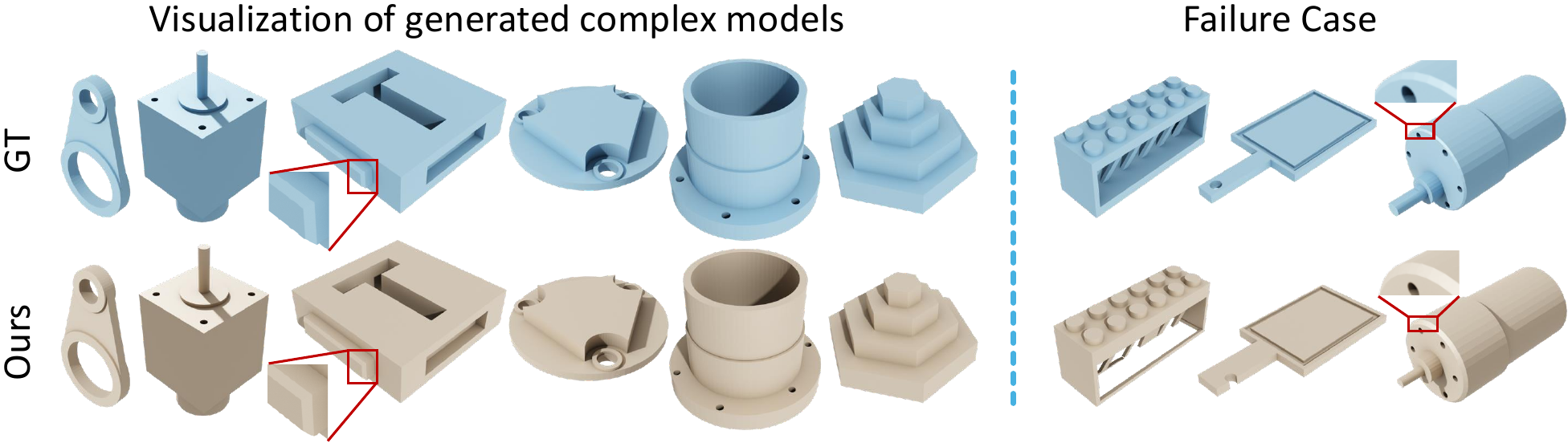}
  \caption{ \small
    \textbf{Showcase of complex CAD model generation.} 
  }
  \label{fig:samples}
\end{figure}

To demonstrate the capabilities and functional boundaries of our method, we visualize a set of generated complex CAD cases in Figure \ref{fig:samples}. Each displayed case involves at least four non-sketch operations, representing multi-step construction and refinement. While our method is capable of generating complex CAD models, it occasionally encounters failures where a specific part is mispositioned relative to the ground truth, leading to slight discrepancies in the final model.

\section{Limitations and Future Work}
Despite Pointer-CAD’s strong performance and structural advantages, several limitations remain and offer promising directions for future research. Although the proposed representation is modality-agnostic by design, our current evaluation focuses primarily on text-conditioned settings. Since real-world CAD workflows often involve multi-modal inputs like images or point clouds, integrating pointer-based generation with robust multi-modal perception remains an open challenge. Furthermore, this work focuses on single-part modeling and does not yet address assembly-level relationships, such as mate constraints and hierarchical dependencies, which are essential for full-scale CAD automation.

\section{Conclusion}
In this work, we present Pointer-CAD, an LLM-driven framework extending command-sequence CAD generation to include \textit{chamfer} and \textit{fillet} operations. To ensure geometric accuracy and enable entity selection, we introduce a pointer-based command sequence incorporating B-rep geometry, allowing the model to reference existing faces and edges.
Experiments demonstrate that Pointer-CAD produces models with higher topological accuracy and geometric fidelity compared to existing text-conditioned methods.

\section{Acknowledgements}
This work is supported by General Research Fund of the Research Grants Council (grant \#17200725), partially supported by the Shenzhen Loop Area Institute, and in part by the JC STEM Lab funded by The Hong Kong Jockey Club Charities Trust.

{
    \small
    \bibliographystyle{ieeenat_fullname}
    \bibliography{main}
}

\newpage
\definecolor{spatialBlue}{RGB}{65, 105, 185}
\definecolor{objectRed}{RGB}{200, 60, 60}

In these supplementary materials, we provide the following:
\begin{itemize}[left=1.5em]
\item Additional evaluations and analytical discussions, including comprehensive metrics for text-to-CAD assessment, comparisons on multiple benchmarks, and ablation studies on invalidity ratio;
\item Details of the proposed pointer-based representation, including sketch plane selection method, specific vector translation rules, and definitions of geometric special cases in pointer-based referencing;
\item Details of the training framework, covering the B-rep encoder, implementation details of the autoregressive decoder, and the training objective;
\item Visualization of some dataset cases, the prompts used for annotation, and dataset statistics;
\item Implementation Details;
\end{itemize}

\section{Additional Evaluations and Analytical Discussions}
\label{sec:more_results}

\subsection{Comprehensive Metrics for Text-to-CAD Evaluation}

To provide a more complete assessment of model performance, we further report the Invalidity Ratio (IR), Dangling Edge Length (DangEL), and Self-Intersection Ratio (SIR) following previous works \cite{xu2024cad,govindarajan2025cadmium}. 
IR measures generation robustness, while DangEL and SIR quantify complementary aspects of topological soundness. Also, following CADFusion~\cite{wang2024cadfusion} and SkexGen~\cite{xu2022skexgen}, we evaluate our approach using the LVM score, Coverage (COV), Minimum Matching Distance (MMD), and Jensen-Shannon Divergence (JSD).

\noindent \textbf{Clarification on computation of IR:} 

Text2CAD~\cite{khan2024text2cad} defines IR as:
\begin{equation}
  \text{IR} = \frac{N_{\text{valid}} - N_{\text{post\_build}}}{N_{\text{valid}}} ,
\end{equation}
where $N_{\text{valid}}$ is the number of representations that become well-formatted \textbf{after post-processing}, and $N_{\text{post\_build}}$ is the subset that can be built into non-zero-volume solids \textbf{under the same post-processing pipeline}.

To eliminate the influence of hardcoded post-processing, we adopt another definition of IR across all methods, given by:
\begin{equation}
  \text{IR} = \frac{N_{\text{test}} - N_{\text{build}}}{N_{\text{test}}} ,
\end{equation}
where $N_{\text{build}}$ denotes the number of generated representations that can be successfully built into non-zero-volume solids \textbf{without any post-processing}, and $N_{\text{test}}$ is the size of the test set.
Under our definition, any output with malformed or incorrect formatting is directly treated as invalid, enforcing a stricter and more realistic failure criterion than the IR used in Text2CAD.

\subsection{Additional results on our dataset}
\label{sec:additional_metric}

Table \ref{tab:performance-deepcad} presents additional evaluation results for various metrics (IR, DangEL, and SIR) on the Recap-DeepCAD dataset.
Pointer-CAD consistently outperforms all baselines, achieving the lowest IR and demonstrating superior topological soundness.
These improvements show that our pointer mechanism enhances robustness and yields more coherent and structurally consistent CAD constructions. 

To further investigate this robustness against error accumulation during sequential generation, Table \ref{tab:step} evaluates single- versus multi-extrusion ($\ge 2$) models on the Recap-DeepCAD dataset. While baseline methods experience significant performance drops in multi-step scenarios, Pointer-CAD maintains exceptional stability. 

Table \ref{tab:performance-omnicad+} further evaluates \textit{chamfer} and \textit{fillet} operations.
Pointer-CAD is trained on the Recap-OmniCAD\textsuperscript{+} dataset, whereas baselines that do not support these operations are trained on Recap-OmniCAD for fairness.
Within this expanded operation space, Pointer-CAD continues to outperform all baselines, indicating that its modeling accuracy and stability are maintained despite the increased operation diversity.

Finally, Table \ref{tab:metrics} presents comprehensive generative evaluation metrics, COV, MMD, JSD, and LVM score, across both the Recap-DeepCAD and Recap-OmniCAD\textsuperscript{+} datasets. Specifically, COV, MMD, and JSD are computed using 1K real and 4K generated samples, averaged over three independent runs. The LVM score is evaluated on the full test set via GPT-4o, utilizing the official evaluation prompt from CADFusion. As demonstrated, our method consistently surpasses all baselines across these metrics on both datasets.

\begin{table}
  \setlength{\tabcolsep}{3pt}
  \caption{
    \textbf{Quantitative comparison on the Recap-DeepCAD dataset across additional metrics.}
    Pointer-CAD consistently achieves lower invalidity and stronger topological soundness than all baseline methods, demonstrating its superior robustness and reliability in generating coherent CAD constructions.
  }
  \label{tab:performance-deepcad}
  \centering
  \begin{adjustbox}{max width=\linewidth}
\begin{tabular}{cccccccc}
\toprule
\multirow{2}[4]{*}{Model} & \multirow{2}[4]{*}{DeepCAD} & \multirow{2}[4]{*}{Text2CAD} & \multicolumn{3}{c}{CADmium} & \multicolumn{2}{c}{Pointer-CAD} \\
\cmidrule{4-8}      &       &       & 1.5B  & 3B    & 7B    & 0.5B  & 1.5B \\
\midrule
IR \textdownarrow & 39.23  & 30.16  & 42.84  & 36.34  & 31.79  & 15.02  & \textbf{8.80 } \\
DangEL \textdownarrow & 1.80  & 0.71  & 3.27  & 3.88  & 5.33  & 0.20  & \textbf{0.19 } \\
SIR \textdownarrow & 0.15  & 0.07  & 0.10  & 0.13  & 0.20  & \textbf{0.02}  & \textbf{0.02 } \\
\bottomrule
\end{tabular}%

  \end{adjustbox}
\end{table}

\begin{table}
  \caption{
    \textbf{Quantitative error accumulation analysis on Recap-DeepCAD.}
    We evaluate single- vs. multi-extrusion ($\ge 2$) models.
    Our method shows minimal performance drop across steps.
    Best results are highlighted in \textcolor{objectRed}{red} (single) and \textcolor{spatialBlue}{blue} (multi).
  }
  \label{tab:step}
  \centering
  \begin{adjustbox}{max width=\linewidth}
\begin{tabular}{ccccccccccccc}
\toprule
\multirow{3}[6]{*}{Model} & \multicolumn{2}{c}{\multirow{2}[4]{*}{DeepCAD}} & \multicolumn{2}{c}{\multirow{2}[4]{*}{Text2CAD}} & \multicolumn{4}{c}{CADmium} & \multicolumn{4}{c}{Pointer-CAD} \\
\cmidrule{6-13}  & \multicolumn{2}{c}{} & \multicolumn{2}{c}{} & \multicolumn{2}{c}{1.5B} & \multicolumn{2}{c}{3B} & \multicolumn{2}{c}{0.5B} & \multicolumn{2}{c}{1.5B} \\
\cmidrule{2-13}  & Single & Multi & Single & Multi & Single & Multi & Single & Multi & Single & Multi & Single & Multi \\
\midrule
Line F1 \textuparrow & 86.22 & 63.02  & 92.54  & 80.62  & 92.94  & 63.51  & 95.08  & 56.57  & 96.54  & 94.06  & \textbf{\textcolor{objectRed}{99.34}}  & \textbf{\textcolor{spatialBlue}{97.65}}  \\
Arc F1 \textuparrow & 41.74 & 11.97  & 56.32  & 33.21  & 28.61  & 9.64  & 41.73  & 8.18  & 70.31  & 65.91  & \textbf{\textcolor{objectRed}{96.52}}  & \textbf{\textcolor{spatialBlue}{93.50}}  \\
Circle F1 \textuparrow & 86.13 & 62.70  & 94.05  & 78.62  & 92.75  & 37.12  & 93.41  & 40.48  & 98.87  & 96.52  & \textbf{\textcolor{objectRed}{99.71}}  & \textbf{\textcolor{spatialBlue}{97.35}}  \\
Extrusion F1 \textuparrow & 95.34 & 83.39  & 99.96  & 95.90  & \textbf{\textcolor{objectRed}{100}}  & 68.18  & \textbf{\textcolor{objectRed}{100}}  & 62.43  & 99.99  & \textbf{\textcolor{spatialBlue}{99.10}}  & \textbf{\textcolor{objectRed}{100}}  & 98.88  \\
CD mean \textdownarrow & 32.48 & 46.52  & 12.82  & 25.95  & 7.75  & 23.53  & 6.79  & 24.25  & 3.80  & 7.39  & \textbf{\textcolor{objectRed}{1.70}}  & \textbf{\textcolor{spatialBlue}{4.35}}  \\
CD median \textdownarrow & 6.92 & 22.76  & 1.14  & 13.24  & 0.34  & 12.46  & 0.26  & 11.54  & 0.33  & 1.76  & \textbf{\textcolor{objectRed}{0.24}}  & \textbf{\textcolor{spatialBlue}{0.63}}  \\
\bottomrule
\end{tabular}%

  \end{adjustbox}
\end{table}

\begin{table}
  \setlength{\tabcolsep}{3pt}
  \caption{
    \textbf{Quantitative evaluation on the Recap-OmniCAD\textsuperscript{+} dataset.}
    Pointer-CAD maintains superior performance over all baselines even under expanded operation diversity.
  }
  \label{tab:performance-omnicad+}
  \centering
  \begin{adjustbox}{max width=\linewidth}
\begin{tabular}{cccccccc}
\toprule
\multirow{2}[4]{*}{Model} & \multirow{2}[4]{*}{DeepCAD} & \multirow{2}[4]{*}{Text2CAD} & \multicolumn{3}{c}{CADmium} & \multicolumn{2}{c}{Pointer-CAD} \\
\cmidrule{4-8}      &       &       & 1.5B  & 3B    & 7B    & 0.5B  & 1.5B \\
\midrule
IR \textdownarrow & 42.71  & 33.72  & 46.65  & 40.17  & 34.38  & 25.37  & \textbf{19.15 } \\
DangEL \textdownarrow & 3.34  & 1.13  & 3.20  & 4.32  & 4.61  & 0.28  & \textbf{0.27 } \\
SIR \textdownarrow & 0.19  & 0.11  & 0.12  & 0.15  & 0.20  & \textbf{0.02 } & \textbf{0.02 } \\
\bottomrule
\end{tabular}%

  \end{adjustbox}
\end{table}

\begin{table}
  \setlength{\tabcolsep}{3pt}
  \caption{
    \textbf{Additional evaluation.}
    Our method consistently outperforms all baselines across these metrics.
  }
  \label{tab:metrics}
  \centering
  \begin{adjustbox}{max width=\linewidth}
\begin{tabular}{ccccccc|cccccc}
\toprule
  & \multicolumn{6}{c|}{(a) Recap-DeepCAD} & \multicolumn{6}{c}{(b) Recap-OmniCAD\textsuperscript{+}} \\
\cmidrule{2-13}\multirow{2}[4]{*}{Model} & \multirow{2}[4]{*}{DeepCAD} & \multirow{2}[4]{*}{Text2CAD} & \multicolumn{2}{c}{CADmium} & \multicolumn{2}{c|}{Pointer-CAD} & \multirow{2}[4]{*}{DeepCAD} & \multirow{2}[4]{*}{Text2CAD} & \multicolumn{2}{c}{CADmium} & \multicolumn{2}{c}{Pointer-CAD} \\
\cmidrule{4-7}\cmidrule{10-13}  &   &   & 1.5B & 3B & 0.5B & 1.5B &   &   & 1.5B & 3B & 0.5B & 1.5B \\
\midrule
COV(\%) \textuparrow & 71.85  & 75.90  & 69.00  & 70.20  & 86.97  & \textbf{89.40 } & 63.68  & 72.22  & 65.70  & 66.67  & \textbf{87.57 } & 87.53  \\
MMD \textdownarrow & 1.29  & 0.97  & 1.20  & 1.19  & 0.77  & \textbf{0.70 } & 1.48  & 1.08  & 1.29  & 1.16  & \textbf{0.74 } & 0.77  \\
JSD (x100) \textdownarrow & 3.12  & 3.04  & 1.45  & 1.82  & 0.84  & \textbf{0.62 } & 4.03  & 3.48  & 1.79  & 2.21  & 0.66  & \textbf{0.65 } \\
LVM Score \textuparrow & 5.41  & 5.82  & 7.28  & 7.52  & 8.46  & \textbf{8.57 } & 5.39  & 5.64  & 6.98  & 7.03  & \textbf{7.99 } & 7.96  \\
\bottomrule
\end{tabular}%

  \end{adjustbox}
\end{table}

\begin{table*}
  \caption{
    \textbf{Quantitative comparison on Text2CAD datasets.}
    Pointer-CAD clearly surpasses Text2CAD and CADmium, and remains highly competitive with Text-to-CadQuery despite the latter’s reliance on the mature CadQuery engine.
  }
  \label{tab:performance-text2cad}
  \centering
  \begin{threeparttable}
\begin{tabular}{ccccccc}
\toprule
\multirow{2}[4]{*}{Model} & \multirow{2}[4]{*}{Text2CAD} & Text-to- & \multicolumn{2}{c}{CADmium} & \multicolumn{2}{c}{Pointer-CAD} \\
\cmidrule{4-7}      &       & CadQuery-1.5B & 1.5B  & 3B    & 0.5B  & 1.5B \\
\midrule
IR \textdownarrow & 9.58  & 12.27  & 9.12  & 8.51  & 6.74  & \textbf{6.59 } \\
Line F1 \textuparrow & 83.78  & -     & 79.00  & 80.30  & 84.01  & \textbf{85.77 } \\
Arc F1 \textuparrow & 41.54  & -     & 37.72  & 44.22  & 37.51  & \textbf{46.83 } \\
Circle F1 \textuparrow & 78.38  & -     & 76.42  & 78.39  & 77.93  & \textbf{78.66 } \\
Extrusion F1 \textuparrow & 93.76  & -     & 88.53  & 88.06  & \textbf{95.72 } & 95.50  \\
CD mean \textdownarrow & 11.78  & 47.72  & 11.69  & 10.83  & 10.17  & \textbf{8.57 } \\
CD median \textdownarrow & 0.33  & 27.75  & 0.44  & 0.37  & 0.33  & \textbf{0.30 } \\
SegE \textdownarrow & 0.41  & 0.24  & 0.88  & 0.91  & \textbf{0.21 } & 0.22  \\
DangEL \textdownarrow & 0.43  & 0.25  & 2.10  & 1.86  & \textbf{0.16 } & 0.17  \\
SIR \textdownarrow & 0.02  & \textbf{0.01 } & 0.07  & 0.06  & 0.02  & 0.02  \\
FluxEE \textdownarrow & 9.40  & \textbf{0.36 } & 6.30  & 4.79  & 2.56  & 2.44  \\
\bottomrule
\end{tabular}%

  \begin{tablenotes}
  \footnotesize
    \item[*] Due to inherent errors in the process by which Text-to-CadQuery constructs its CadQuery dataset, the ground-truth CadQuery models naturally deviate from the original Text2CAD models. When computing the error between the predicted models and these converted ground-truth models, we obtain \textbf{a mean Chamfer Distance of 12.79 and a median of 0.32.} 
  \end{tablenotes}
  \end{threeparttable}
\end{table*}

\subsection{Results on the Text2CAD Dataset}

In addition to the baselines used in the main paper, we also include Text-to-CadQuery~\cite{xie2025text}, which represents CAD models using CadQuery~\cite{cadquery} code. We compare our proposed method with existing approaches on the dataset proposed in Text2CAD, whose annotations are parameter-normalized and unit-free, resulting in a simpler annotation format compared with our Recap-DeepCAD and Recap-OmniCAD\textsuperscript{+} datasets, as shown in Figure \ref{fig:norm_prompt_comparison}. 

Note that, Text-to-CadQuery constructs its CadQuery dataset sourced from Text2CAD dataset and converts the original Text2CAD models into CadQuery scripts format.
However, this conversion procedure introduces non-negligible geometric discrepancies, causing the reconstructed ground truth CadQuery models to deviate from the original Text2CAD geometry. Therefore, for Text-to-CadQuery, we additionally report the Chamfer distance between its predicted models and the converted ground-truth models, as noted in the footnote. As reported in Table \ref{tab:performance-text2cad}, Pointer-CAD-1.5B achieves the best IR, sketch operation F1, and CD, while Pointer-CAD-0.5B attains superior extrusion F1, SegE, and DangEL.

\noindent \textbf{Comparison with Text-to-CadQuery.} Comparing the results of Pointer-CAD and Text-to-CadQuery, we observe that Text-to-CadQuery achieves lower SIR and FluxEE. This improved performance stems from its use of the mature CadQuery engine and its built-in post-processing, which inherently produces geometrically valid outputs. For instance, the following code directly creates a box without self-intersections while ensuring watertightness.

\begin{verbatim}
result = Workplane("front").box(
    width, width, thickness
)
\end{verbatim}

However, Pointer-CAD achieves smaller SegE and DangEL, even though Text-to-CadQuery relies on a highly engineered drawing pipeline designed to prevent discontinuities and connectivity errors.
These results collectively demonstrate that the pointer mechanism in Pointer-CAD produces more coherent and well-structured CAD operations, enabling stronger geometric consistency than both sequence-based and engine-assisted baselines.

\subsection{Ablation Study on Invalidity Ratio}

\begin{figure*}
  \centering
  \input{figures/norm_prompt_comparison}
  \caption{
    \textbf{Prompt comparison.}
    Recap-DeepCAD dataset includes dimensional values with explicit units, whereas Text2CAD dataset uses normalized, unit-free geometric parameters.
  }
  \label{fig:norm_prompt_comparison}
\end{figure*}

\begin{table}
  \setlength{\tabcolsep}{3pt}
  \caption{
    \textbf{Ablation results on the Recap-DeepCAD-Norm dataset.}
    All baseline methods show a substantial drop in IR, indicating that they depend on memorizing dataset-specific dimensional patterns rather than engaging in genuine geometric reasoning.
  }
  \label{tab:performance-deepcad-}
  \centering
  \begin{adjustbox}{max width=\linewidth}
\begin{tabular}{cccc}
\toprule
Model & Normed & Truncate & IR \\
\midrule
\multirow{3}[2]{*}{Text2CAD} & \xmark & \xmark & 30.16  \\
      & \cmark & \xmark & 15.85  \\
      & \xmark & \cmark & 24.41  \\
\midrule
\multirow{2}[2]{*}{Pointer-CAD-0.5B} & \xmark & \xmark & 15.02  \\
      & \cmark & \xmark & 6.13  \\
\midrule
\multirow{2}[2]{*}{Pointer-CAD-1.5B} & \xmark & \xmark & 8.80  \\
      & \cmark & \xmark & 5.37  \\
\bottomrule
\end{tabular}%

  \end{adjustbox}
\end{table}

To investigate the factors contributing to the observed IR disparity between models trained on the Recap-DeepCAD and Text2CAD datasets, we first compare the annotation pipelines used in each dataset. We note that the main difference lies in the presence of units and whether geometric parameters are normalized.
As shown in Figure \ref{fig:norm_prompt_comparison}, the annotations in Text2CAD recenter each model at the origin and uniformly scale it to fit within a canonical cube, effectively pre-aligning its position and size to match the final representation, whereas Recap-DeepCAD annotations do not apply any such simplification.
We argue that this normalization reduces the difficulty of the task. To validate this, we construct a modified Recap-DeepCAD-Norm dataset by removing all units and normalizing all geometric parameters in the original annotations, following the pipeline used in Text2CAD.

As illustrated in Table \ref{tab:performance-deepcad-}, all baseline methods trained on these normalized prompts exhibit a substantial reduction in IR, demonstrating that the complex and heterogeneous dimension annotations in Recap-DeepCAD are a primary source of invalid generations. We also note that Text2CAD has a maximum token limit of 512; therefore, for Text2CAD, we specifically filtered out models exceeding this length and then evaluated the remaining IR again. We also observed a reduction in IR after this filtering.

\subsection{Quantification of Quantization Error}
\begin{figure}[thb]
  \centering
  \includegraphics[width=0.7\linewidth]{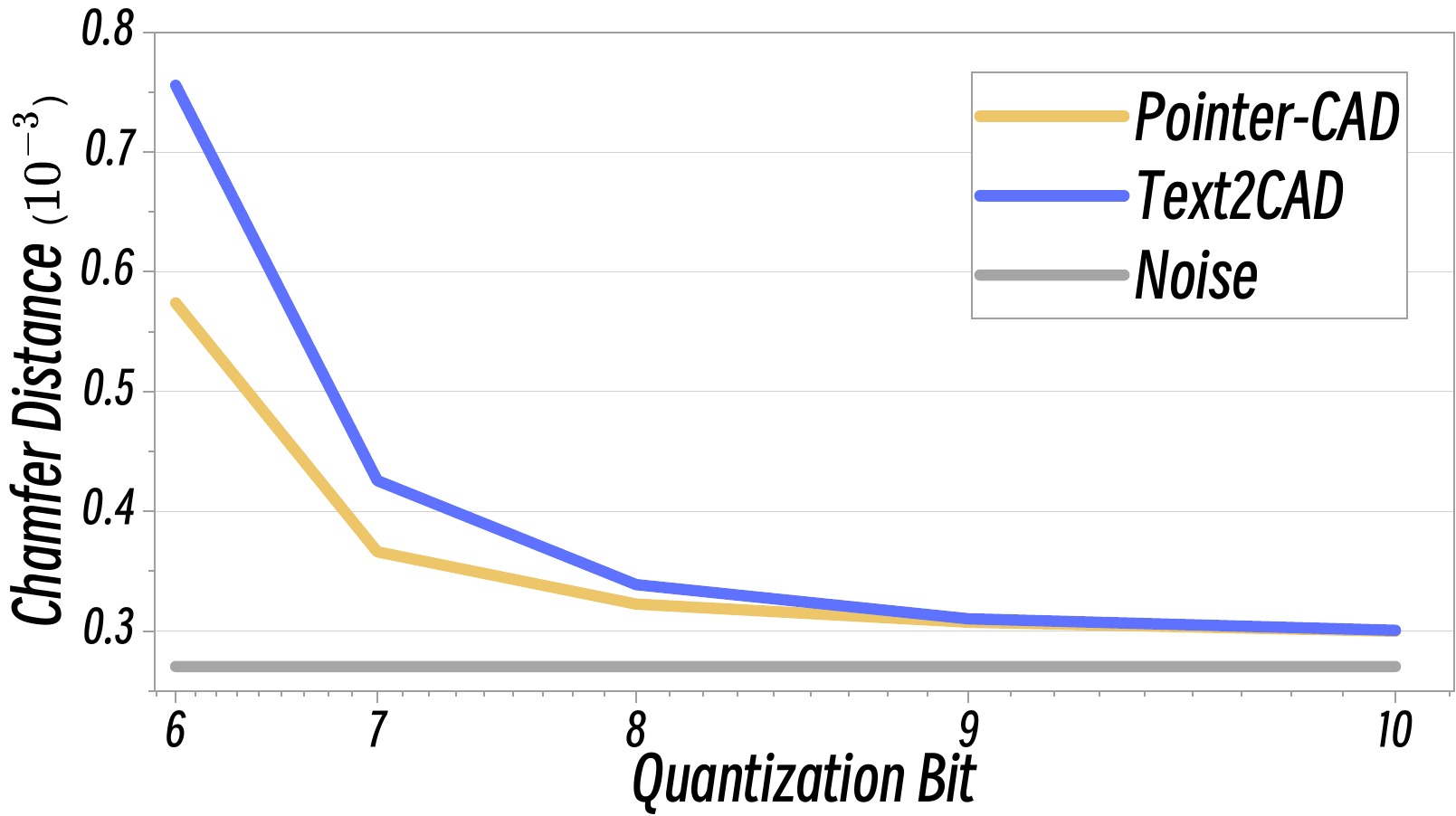}

  \caption{
    \textbf{Quantization Error.}
    We directly measure quantization error by computing the median Chamfer Distance between each representation before and after quantization, where Pointer-CAD exhibits substantially smaller error than Text2CAD.
  }
  \label{fig:quantization_error}
\end{figure}

As discussed in Section 3 of the main paper, command sequences discretize continuous parameters (e.g., coordinates, extrusion distances, angles) into $2^q$ levels. By default, $q=8$ in our experiment.
To assess the reconstruction error introduced by different quantization granularities and different representations, we evaluate the median Chamfer Distance between the ground-truth mesh and the mesh reconstructed from its ground-truth command sequence after applying quantization at various bit widths. A lower Chamfer Distance indicates that the representation preserves geometric fidelity more effectively under quantization. As shown in Figure \ref{fig:quantization_error}, Pointer-CAD consistently exhibits lower quantization error than Text2CAD across all settings.
Moreover, as the quantization bit width increases, the gap between the two methods gradually narrows, eventually approaching toward the inherent CD noise floor induced by random point sampling.

\subsection{Application of Click Interaction Editing}

\begin{figure}[t]
  \centering
  \includegraphics[width=0.7\linewidth]{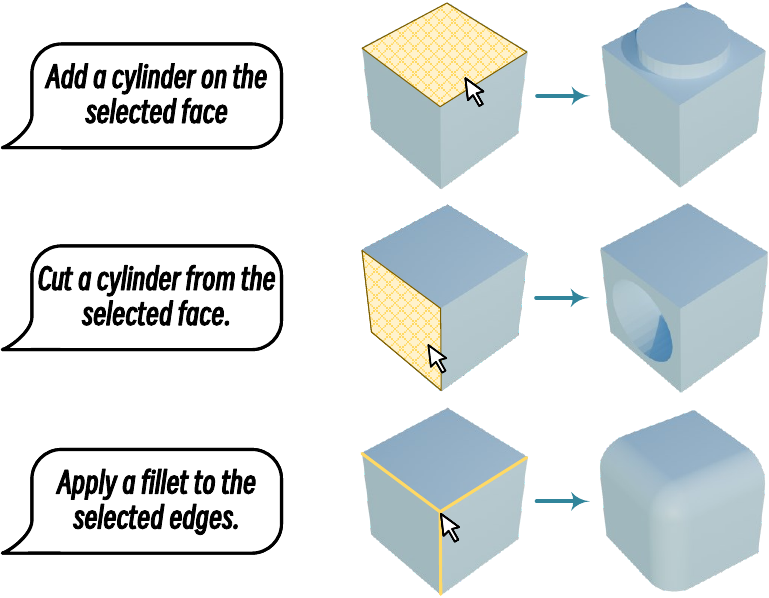}
  \caption{ \textbf{Illustration of our interactive editing functionality.}
    Users can directly click on a face or edge of the CAD model and provide a text prompt to specify the desired operation. 
  }
  \label{fig:application}
\end{figure}

Since our proposed pointer-based command sequence allows entity selection at each step, we extend the model with token concatenation to incorporate user-interactive selections alongside text instructions, enabling an immersive editing experience.
As illustrated in Figure \ref{fig:application}, users can interactively select faces or edges on the current B-rep to explicitly specify the operation target, enabling more precise and intuitive editing through direct manipulation in conjunction with text instructions.

\section{Details of the Pointer-based Representation}
\label{sec:pointer_details}
This section elaborates on the implementation logic of the pointer-based representation and the methodology for sketch plane selection.

\subsection{Specific Vector Translation Rules}
\label{sec:translate_details}
Each token is classified as one of three types: \textit{Label Token}, \textit{Value Token}, or \textit{Pointer}.
To simplify the model architecture, we assign non-overlapping integer ranges to label and value tokens, allowing them to be decoded by a single prediction head.
However, since a pointer is a reference to a geometric entity rather than a simple value, it requires a separate prediction head for decoding.
To distinguish pointers from label and value tokens, we reserve two specific integer values within the label/value token space.
When the model predicts one of these integers, it signals that the current token is a pointer.
These two integers also represent the pointer's state: as shown in Table \ref{tab:token-notation}, \commandnotation{pe} indicates an enabled pointer that references an edge or face, whereas \commandnotation{pd} signifies a disabled (inactive) pointer.
Specifically, for \commandnotation{nv} and \commandnotation{ag}, we first normalize all continuous parameters to the expected range and then quantize into $2^q$ levels and express them using $q$-bit integers.

\begin{table*}[htb]
  \caption{
    \textbf{Special Token Definitions.}
    This table provides a comprehensive list of all Special Tokens used in our command sequence representation, along with their semantic descriptions.
  }
  \label{tab:token-notation}
  \centering
\begin{tabular}{cccc}
\toprule
Notation & ID   & Type & Description \\
\midrule
\commandnotation{em} & 1    & Label Token & End of model (this step is the final step of the model) \\
\commandnotation{es} & 2    & Label Token & End of step (additional steps are required after this one) \\
\commandnotation{ss} & 3    & Label Token & Start of sketch \\
\commandnotation{se} & 4    & Label Token & Start of extrusion \\
\commandnotation{sc} & 5    & Label Token & Start of chamfer \\
\commandnotation{sf} & 6    & Label Token & Start of fillet \\
\commandnotation{sp} & 7    & Label Token & Start of profile \\
\commandnotation{sl} & 8    & Label Token & Start of loop \\
\commandnotation{sx} & 9    & Label Token & Start of curve \\
\commandnotation{pe} & 10   & Pointer & Pointer to an Edge or Face \\
\commandnotation{pd} & 11   & Pointer & Empty pointer \\
\commandnotation{or} & \{12, 13\} & Label Token & Orientation (Clockwise, Counter-Clockwise) \\
\commandnotation{dr} & $[14,\, 19]$ & Label Token & Direction (X+, X-, Y+, Y-, Z+, Z-) \\
\commandnotation{bo} & $[20,\, 23]$ & Label Token & Boolean (New, Join, Cut, Intersect) \\
\commandnotation{nv} & $[24,\, 24+2^{q})$ & Value Token & Normalized to $[0,\, 1]$ and then quantize to $2^{q}$ level\\
\commandnotation{ag} & $[24,\, 24+2^{q})$ & Value Token & Normalized to angle $[0^\circ,360^\circ)$ and then quantize to $2^{q}$ level\\
\bottomrule
\end{tabular}%

\end{table*}

Based on the notation in Table \ref{tab:token-notation}, we define the translation rules for each command as shown in Table \ref{tab:sequence}.
A CAD model is represented as a sequence of valid sequences, with only the last valid sequence is end with \commandnotation{em}.

\begin{table*}[htb]
  \caption{
    \textbf{Sequence definitions.}
    Each sequence is defined by a specific combination of commands.
    The superscript [ \textsuperscript{+} ] denotes that the element appears one or more times, while the symbol [ / ] indicates that one of the alternatives can be chosen.
  }
  \label{tab:sequence}
  \centering
\begin{tabular}{ccc}
\toprule
Notation  & Sequence & Description \\
\midrule
{\sequencenotation{P}} & \commandnotation{nv}\ \commandnotation{nv}\ \ \commandnotation{pe / pd} & 2D point $(x,y)$, snapped to reference or placed freely \\
\midrule
{\sequencenotation{L}} & \commandnotation{sx}\ \ \sequencenotation{P} & Line starting from a 2D point $(x,y)$ \\
{\sequencenotation{C}} & \commandnotation{sx}\ \ \sequencenotation{P}\ \ \commandnotation{nv} & Circle with center at $(x,y)$ and radius $r$ \\
{\sequencenotation{A}} & \commandnotation{sx}\ \ \sequencenotation{P}\ \ \commandnotation{ag}\ \commandnotation{or} & Arc starting from $(x,y)$ with angle $\alpha$ and orientation \\
\midrule
{\sequencenotation{Loop}} & \commandnotation{sl}\ \ \sequencenotation{L / C / A}\textsuperscript{+} & Closed loop composed of multiple curves \\
{\sequencenotation{Profile}} & \commandnotation{sp}\ \ \sequencenotation{Loop}\textsuperscript{+} & 2D Region defined by one or more loops \\
{\sequencenotation{CS}} & \commandnotation{dr}\ \ \sequencenotation{P}\ \ \commandnotation{ag}\ \commandnotation{nv} & 2D coordinate system in 3D space \\
{\sequencenotation{Sketch}} & \commandnotation{ss}\ \commandnotation{pe}\ \ \sequencenotation{CS}\ \ \sequencenotation{Profile}\textsuperscript{+} & Sketch on a plane specified by pointer \\
\midrule
{\sequencenotation{Extrude}} & \commandnotation{se}\ \ \commandnotation{nv}\ \ \commandnotation{nv}\ \ \commandnotation{bo} & Extrude operation with depth and Boolean type \\
{\sequencenotation{EPart}} & \sequencenotation{Sketch}\textsuperscript{+}\ \ \sequencenotation{Extrude} & Solid part constructed by extrusion \\
\midrule
{\sequencenotation{Chamfer}} & \commandnotation{sc}\ \ \commandnotation{nv}\ \ \commandnotation{pe}\textsuperscript{+} & Chamfer operation on referenced edges \\
{\sequencenotation{Fillet}} & \commandnotation{sf}\ \ \commandnotation{nv}\ \ \commandnotation{pe}\textsuperscript{+} & Fillet operation on referenced edges \\
\midrule
\textbf{{\sequencenotation{VSeq}}} & \textbf{\sequencenotation{EPart / Chamfer / Fillet}\ \ \commandnotation{es / em}} & \textbf{Valid sequence} \\
\bottomrule
\end{tabular}%

\end{table*}

\subsection{Sketch Plane Selection}
\label{sec:plane_select_details}

\begin{figure}[htb]
    \centering
    \begin{subfigure}[t]{0.3\linewidth}
        \centering
        \includegraphics[width=\linewidth]{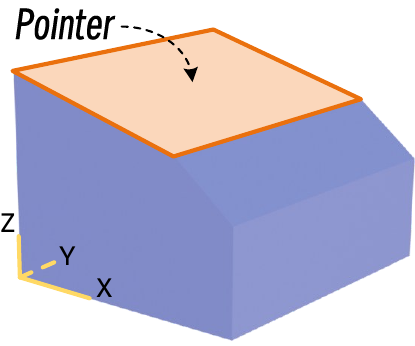}
        \caption{Face selection.}
        \label{fig:coordinate_system_a}
    \end{subfigure}
    \hfill
    \begin{subfigure}[t]{0.3\linewidth}
        \centering
        \includegraphics[width=\linewidth]{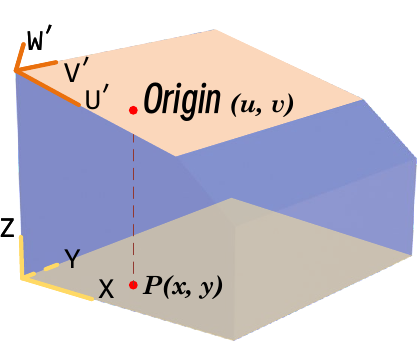}
        \caption{Origin definition.}
        \label{fig:coordinate_system_b}
    \end{subfigure}
    \hfill
    \begin{subfigure}[t]{0.3\linewidth}
        \centering
        \includegraphics[width=\linewidth]{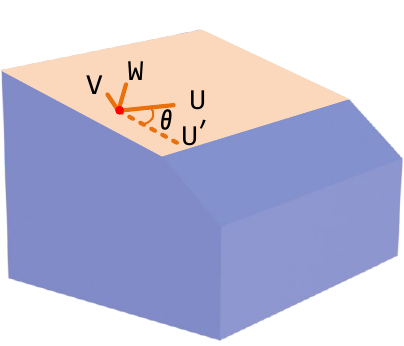}
        \caption{Rotation definition.}
        \label{fig:coordinate_system_c}
    \end{subfigure}
    \caption{
        \textbf{Sketch coordinate system construction.}
        The sketch plane, axes, origin, and rotation are defined step by step to form the local coordinate system $UVW$.
    }
    \label{fig:coordinate_system}
\end{figure}

As defined in the {\sequencenotation{Sketch}} notation, a sketch plane is specified by a \textit{Pointer} to a face and a 2D coordinate system {\sequencenotation{CS}}.
The construction process, illustrated in Figure \ref{fig:coordinate_system}, unfolds in three main steps:
First, a base plane is established by selecting a face with the \textit{Pointer}, as shown in Figure \ref{fig:coordinate_system_a}.
The resulting sketch plane is coplanar with this face.
Second, a local coordinate system $U'V'W'$ is constructed on this plane.
The normal axis, $W'$, is aligned with the face normal that has a positive dot product with a world axis direction, $n$, specified by the \textit{Label Token} \commandnotation{dr}.
The primary in-plane axis, $U'$, is determined by projecting an auxiliary direction, $d$ (listed in Table \ref{tab:direction}), onto the sketch plane.
The second in-plane axis, $V'$, is then derived using the right-hand rule, completing the orthogonal basis $U'V'W'$.
As depicted in Figure \ref{fig:coordinate_system_b}, the origin of this system is defined by projecting a point $P$ from a world coordinate plane onto the sketch plane along the direction $n$.
Finally, as shown in Figure \ref{fig:coordinate_system_c}, the final sketch coordinate system $UVW$ is obtained by applying a counterclockwise in-plane rotation to $U'V'W'$ about the $W$-axis. An optional scaling factor may also be applied to mitigate quantization errors.

\begin{table}[htb]
  \caption{
    \textbf{Direction mapping.}
    In command \commandnotation{dr}, each symbol corresponds to a primary direction and its auxiliary direction.
  }
  \label{tab:direction}
  \centering
\begin{tabular}{c|cc}
\toprule
Symbol & Direction & Auxiliary Direction \\
\midrule
14    & X+    & Y+ \\
15    & X-    & Z+ \\
16    & Y+    & Z+ \\
17    & Y-    & X+ \\
18    & Z+    & X+ \\
19    & Z-    & Y+ \\
\bottomrule
\end{tabular}%

\end{table}

\subsection{Geometric Special Cases in Pointer Referencing}
\label{sec:special_case_details}

While a pointer is generally intended to reference a single, unique geometric entity (i.e., an edge or a face), this one-to-one correspondence breaks down in certain "geometric special cases."
These cases occur when multiple entities are geometrically equivalent from a modeling standpoint, such as coplanar faces or collinear edges.
In such scenarios, selecting any one of these equivalent entities would result in the same final geometry.
Therefore, the ground truth for a pointer is not a single object but rather a set of valid candidates. This section provides precise definitions for these geometric special cases.

\paragraph{Coplanar-Adjacent Faces.}
A face pointer selects a base face to define a sketch plane.
If two or more faces are coplanar (i.e., they lie on the same geometric plane), selecting any of them will result in the same sketch plane definition. Therefore, all faces within such a coplanar group are considered valid candidates for the face pointer.

\paragraph{Collinear-Connected Edges.}
Snapping a sketch point to an existing edge requires an edge pointer.
If other edges are collinear with the target edge, pointing to any of them will produce the same snapping result.
Therefore, all edges within such a collinear group are considered valid candidates for the edge pointer.

\subsection{Pointer Failure Scenarios.} 
\begin{figure}[t]
  \centering
  \includegraphics[width=0.7\linewidth]{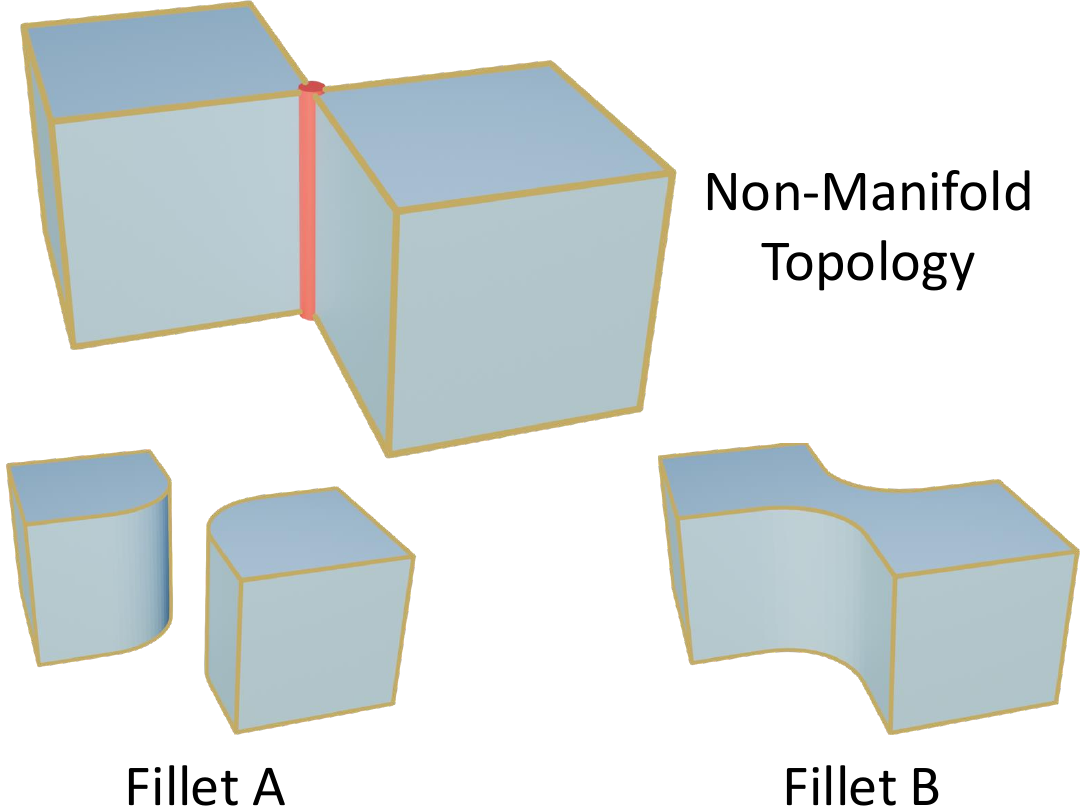}
  \caption{
    A non-manifold topology leads to multiple valid interpretations of a fillet operation.
  }
  \label{fig:pointer_failure}
\end{figure}

The pointer mechanism successfully resolves geometric references in the vast majority of standard modeling scenarios. Failures are largely confined to extreme cases. Primarily, errors arise from inherent limitations within the B-Rep graph when processing non-manifold topologies—specifically, edges bounded by more than two faces. Although such configurations are typically excluded under standard CAD modeling conventions, their presence introduces significant ambiguity in face selection for operations like chamfer and fillet. As illustrated in Figure~\ref{fig:pointer_failure}, encountering a non-manifold edge leads to multiple valid interpretations of a fillet operation, thereby confounding the pointer mechanism.

\section{Details of the training framework.}

\subsection{B-rep encoder}
\label{sec:brep_enc_details}

For each B-rep edge, we uniformly sample 32 points along its parametric curve in 3D space and extract four quantities at each location: point coordinates, tangent and its reverse vector, and first-order derivatives.
Each is represented as a 3D vector, and their concatenation yields a 12-dimensional feature per sample.
Collecting all samples forms an edge feature tensor of shape $32 \times 12$, which serves as input for edge embedding.

For each B-rep face, we uniformly sample its parametric $(u,v)$ domain to construct a regular UV grid of size $32 \times 32$.
At each grid point, we compute the 3D coordinates, unit surface normal, Gaussian curvature, and a binary visibility mask (set to $1$ for interior or boundary samples and $0$ otherwise).
Concatenating these quantities channel-wise gives an 8-dimensional feature per location, i.e., $3+3+1+1$, producing a face tensor of shape $32 \times 32 \times 8$.

For GNN node embeddings (the B-rep face embeddings), we first apply a 2D convolution to the face tensor to expand it to $256$ channels, followed by global adaptive average pooling and a linear projection to a 128-dimensional vector, denoted $h_i^{(0)}$.
Similarly, edge embeddings are obtained by applying 1D convolutions to the edge tensor, expanding it to $256$ channels, followed by global adaptive average pooling and a linear projection to a 128-dimensional vector, denoted $h_{ij}^{(0)}$.
Thus, the graph $\mathcal{G}$ is initialized with node features ${h_i^{(0)}}$ and edge features ${h_{ij}^{(0)}}$ for downstream processing.

\subsection{Implementation Details of the Autoregressive Decoder}

To translate the output of the LLM into our defined command sequence, we process the last hidden state from the model's transformer decoder at each autoregressive decoding step.
We employ a dual-head architecture to decode the hidden state into the appropriate token type.

The first head, which we refer to as the \textbf{Label/Value Head}, is a linear layer responsible for predicting both \textit{Label Tokens} and \textit{Value Tokens}. Its output comprises three parts, consistent with the tokenization scheme introduced earlier:

\begin{itemize}
\item A component corresponding to one of the Label Tokens defined in Table~\ref{tab:token-notation}.
\item A component selecting from two special tokens, \commandnotation{pe} and \commandnotation{pd}, which indicate the pointer state. When the model predicts the \commandnotation{pe}, it indicates that the current token is a pointer, and the output from the second head should be used.
\item A component that corresponds to the quantized bins used for continuous Value Tokens, including those for \commandnotation{nv} or \commandnotation{ag}.
\end{itemize}

The second head, the \textbf{Pointer Head}, is another linear layer specifically designed for decoding pointers. This head's output is a 128-dimensional vector. When the Label/Value Head predicts an active pointer state, this 128-dimensional vector is used to perform a similarity search (via cosine similarity) against the 128-dimensional embeddings of all candidate geometric entities (faces and edges) generated by the B-rep encoder. The entity with the highest similarity score is selected as the pointer's reference. This mechanism allows the model to dynamically ground its generation in the existing B-rep geometry.

\subsection{Details of Training Objective}
\label{sec:training_details}

\paragraph{Pointer Prediction.}
Following CLIP \cite{radford2021learning}, we employ a learnable temperature parameter $\tau$ to control the scale of the logits in the loss computation.
The parameter is initialized to $0.07$, following \cite{wu2018unsupervised}.
To improve training stability, we reparameterize $\tau$ as its reciprocal $s = 1/\tau$ and optimize $\log s$ during training, with $s$ clipped to $s \leq 100$ to avoid excessive scaling of the logits.
The learning rate for $s$ is set to $lr_s = 0.1 \times lr$, and weight decay is not applied during its optimization. 

\paragraph{Overall Objective.}
The overall training objective $\mathcal{L}$ combines the cross-entropy loss for label/value tokens ($\mathcal{L}_v$) and the contrastive loss for pointer tokens ($\mathcal{L}_p$). The final loss is a weighted sum of these two components, controlled by hyperparameters $\lambda_v$ and $\lambda_p$.
In all our experiments, we set $\lambda_v = 0.5$ and $\lambda_p = 0.5$ to give them equal weight.

\section{Details of the Dataset}
\label{sec:dataset_details}

\subsection{Dataset Visualization}

\begin{figure}[t]
  \centering
  \includegraphics[width=\linewidth]{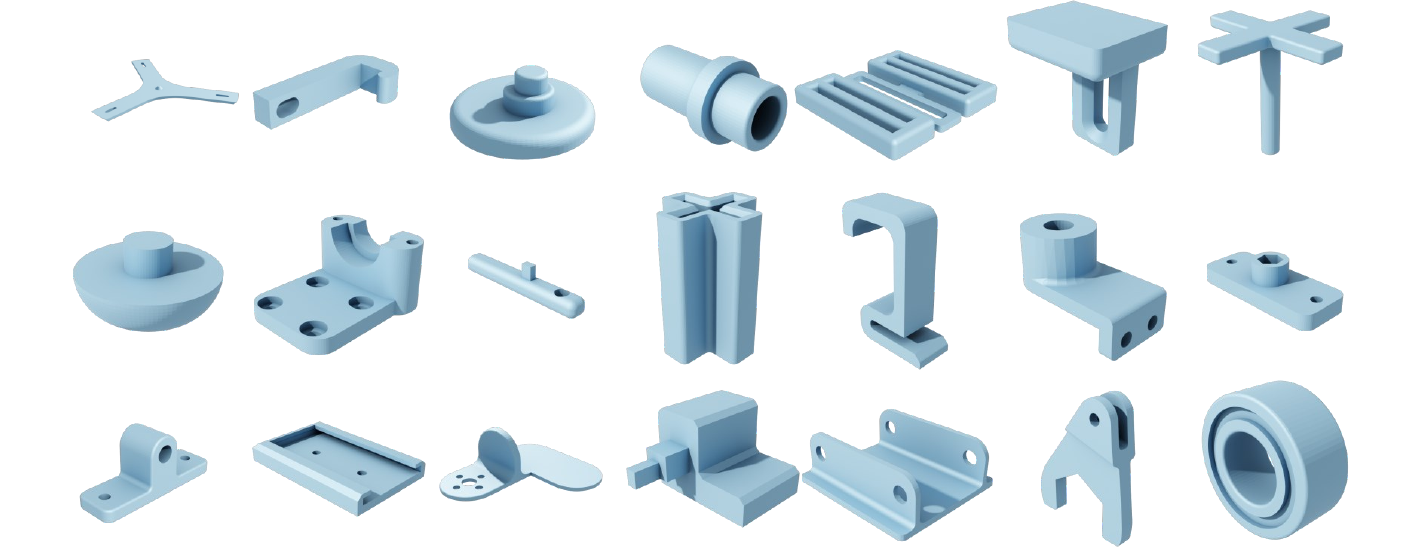}

  \caption{
    \textbf{Representative samples from the Recap-OmniCAD\textsuperscript{+} dataset.}
    The figure displays a range of models with varying complexity, from simpler parts with basic features to intricate components incorporating numerous fillets, chamfers, and complex sketches.
  }
  \label{fig:dataset_visualization}
\end{figure}

Figure \ref{fig:dataset_visualization} presents several representative samples from the Recap-OmniCAD\textsuperscript{+} dataset, showcasing a wide spectrum of model complexity and diversity.
As illustrated, our dataset contains a rich variety of models that not only feature complex geometric details such as fillets and chamfers but also exhibit diverse topological structures like holes, pockets, and multi-body components.

\subsection{Details of Annotation Prompts}
\label{sec:supp_prompt}
\begin{figure*}
  \centering
  \includegraphics[width=0.7\linewidth]{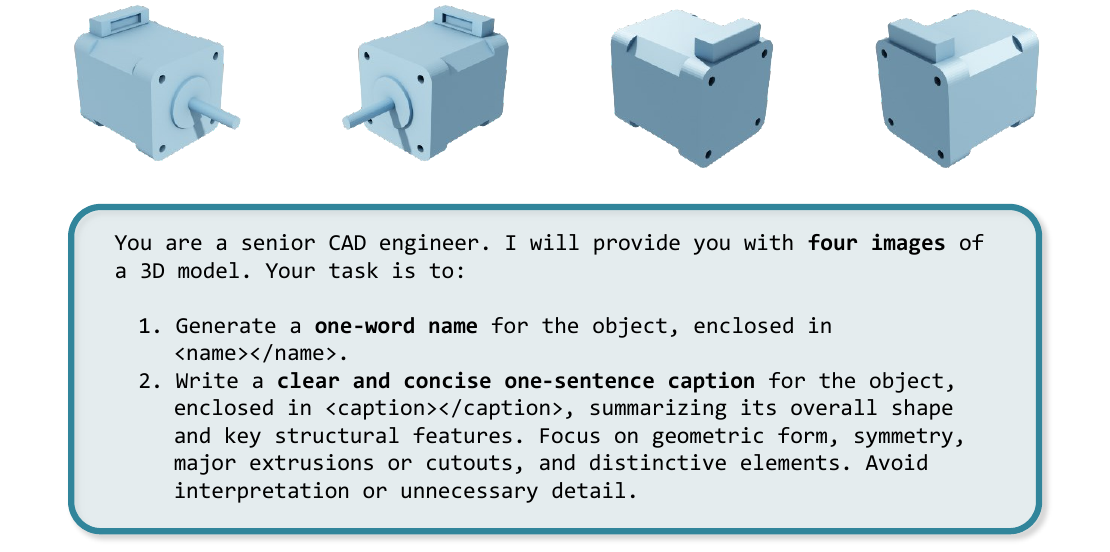}
  \caption{
    \textbf{Prompt for visual description.}
    This prompt is used with the Qwen2.5-vl-72B model to generate a description of the CAD model's visual appearance.
  }
  \label{fig:annotation_vl}
\end{figure*}

\begin{figure*}
  \centering
  \includegraphics[width=0.7\linewidth]{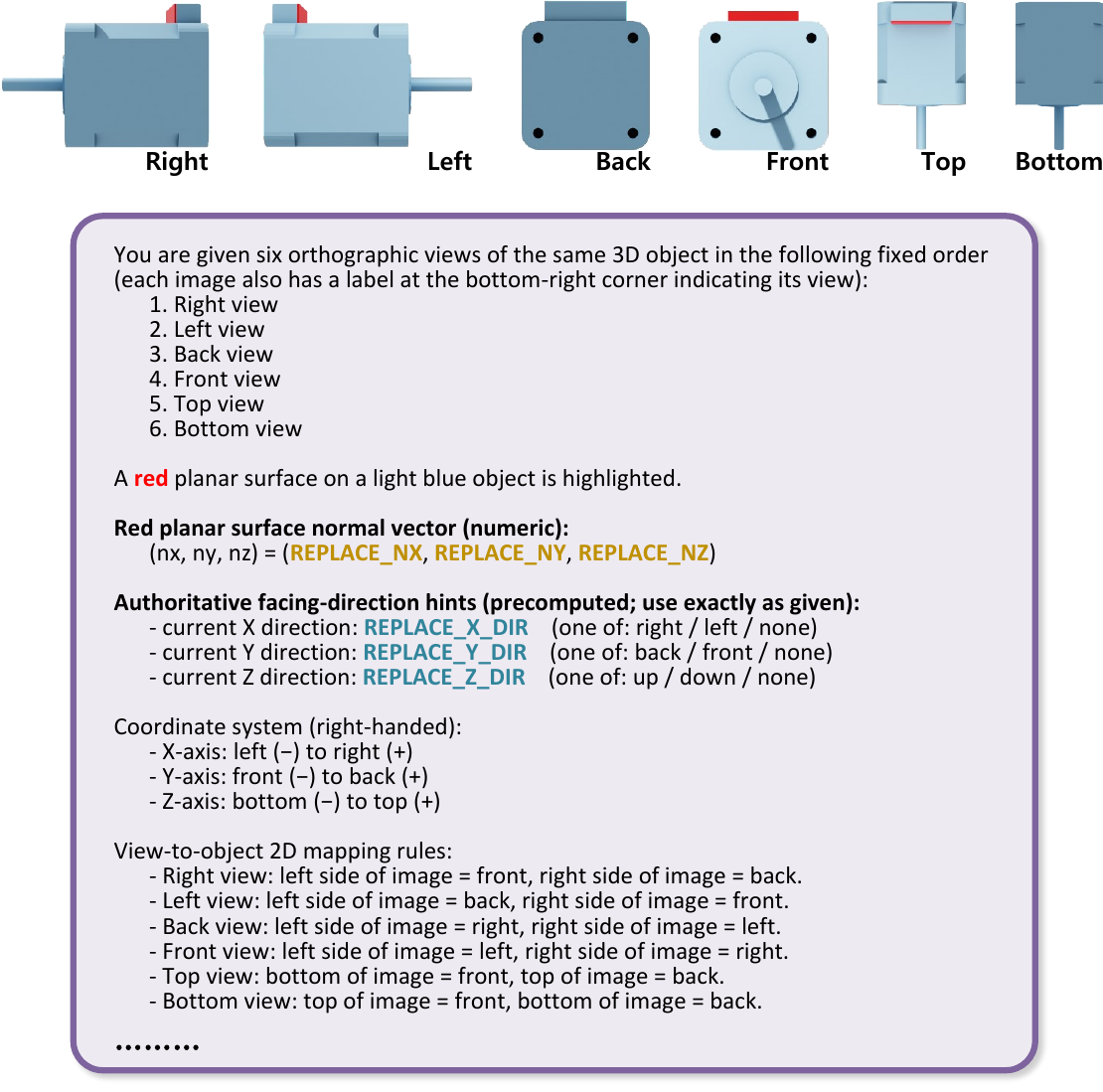}
  \caption{
    \textbf{Prompt for sketch plane description.}
    This prompt guides the model to describe the relative position of the sketch plane, with placeholders for the normal vector and facing direction being dynamically replaced.
  }
  \label{fig:annotation_sp}
\end{figure*}

In our dataset construction process, we employ a multi-step approach to generate rich and detailed annotations for each CAD model.
First, we utilize the Qwen2.5-vl-72B model to generate a visual description of the model's appearance, using the prompt shown in Figure \ref{fig:annotation_vl}.
Next, we use the same model to describe the relative position of the sketch plane within the model, guided by the prompt in Figure \ref{fig:annotation_sp}.
To ensure a clear and accurate understanding for the model, we dynamically replace the placeholders in the prompt with the actual sketch plane surface normal vector and facing direction for each CAD model.

\begin{figure*}
  \centering
  \includegraphics[width=\linewidth]{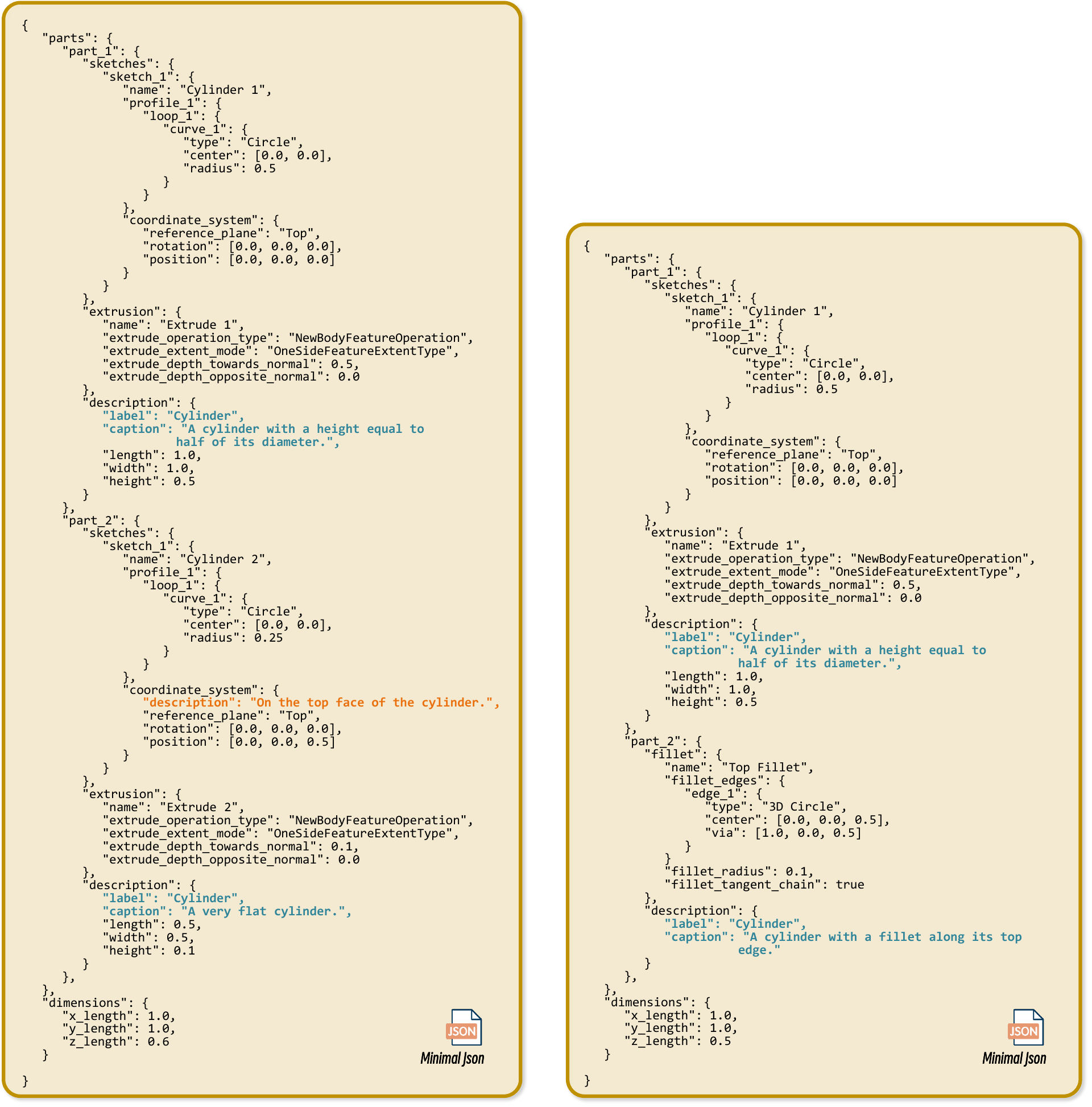}
  \caption{
    \textbf{Examples of the minimal JSON structure.}
    This figure illustrates two structured 'minimal JSONs' format, which integrates visual annotations and key modeling parameters for the language model.
  }
  \label{fig:annotation_json}
\end{figure*}

The resulting annotations are then combined with modeling parameters extracted from the raw JSON file to create a structured "minimal JSON," as illustrated in Figure \ref{fig:annotation_json}.
This minimal JSON, along with the prompt shown in Figure \ref{fig:annotation_llm}, is then passed to Qwen2.5-72B-Instruct to generate a final natural language description of the modeling process.

\subsection{Dataset Statistics}
\begin{figure}[t]
  \centering
  \includegraphics[width=0.8\linewidth]{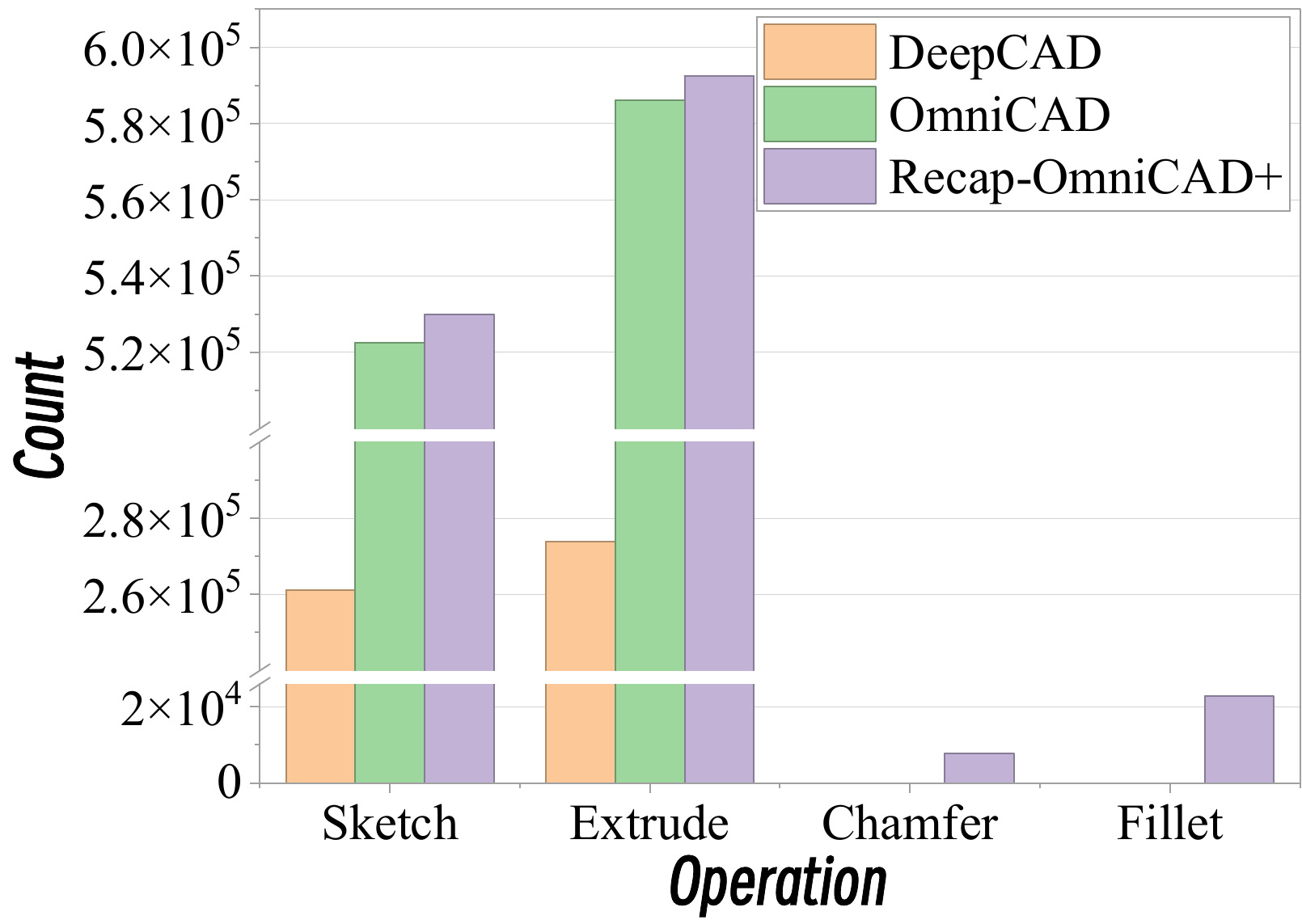}

  \caption{
    \textbf{Distribution of modeling operations across datasets.}
    The figure illustrates the total count of each modeling operation type for the DeepCAD, OmniCAD, and Recap-OmniCAD\textsuperscript{+} datasets.
  }
  \label{fig:dataset_statistics_operation}
\end{figure}

\begin{figure}[t]
  \centering
  \includegraphics[width=0.8\linewidth]{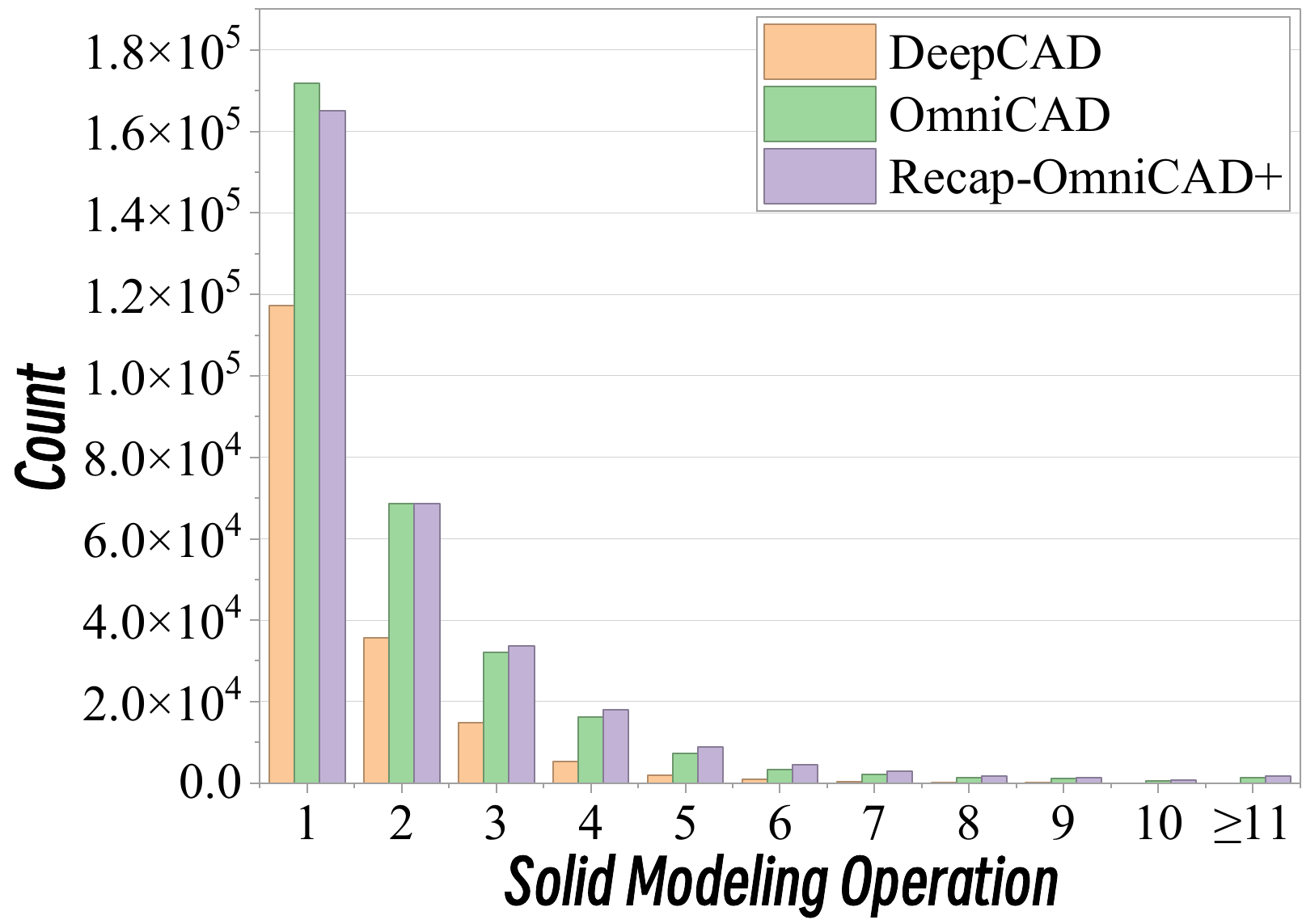}

  \caption{
    \textbf{Distribution of modeling steps per model.}
    The figure compares the number of solid modeling operations required per model across the datasets.
  }
  \label{fig:dataset_statistics_step}
\end{figure}

We provide a statistical analysis of our dataset in Figure \ref{fig:dataset_statistics_operation} and Figure \ref{fig:dataset_statistics_step}.

Figure \ref{fig:dataset_statistics_operation} illustrates the distribution of modeling operations.
Notably, Recap-OmniCAD\textsuperscript{+} includes \textit{chamfer} and \textit{fillet} operations, which are absent in the original OmniCAD.
The reintegration of these features results in a higher count for all operation types in Recap-OmniCAD\textsuperscript{+} compared to OmniCAD.

In Pointer-CAD, the command sequence of a complete CAD model is decomposed into three types of operations: sketch–extrude combinations, chamfers, and fillets. Figure  \ref{fig:dataset_statistics_step} presents the statistics of our dataset according to this decomposition.
The inclusion of \textit{chamfer} and \textit{fillet} operations increases the overall complexity and the average number of steps required to construct a model.
This is reflected in the distribution, where Recap-OmniCAD\textsuperscript{+} has a slightly lower count of models with a single operation but a consistently higher count for models requiring more than one operation compared to OmniCAD.

Furthermore, both figures highlight that OmniCAD and Recap-OmniCAD\textsuperscript{+} are significantly more complex than DeepCAD.
They feature a greater total number of operations and a higher proportion of models requiring a large number of construction steps.
This demonstrates that our datasets are more challenging and better reflect the complexity of real-world CAD modeling tasks.

\section{Implementation Details}
\label{sec:supp_implement}
For the default 0.5B model setting, the entire training process requires approximately 23 hours on 16 NVIDIA H800 GPUs.
We use the AdamW optimizer \citep{loshchilov2017decoupled} with a learning rate of $1 \times 10^{-4}$ and a linear decay schedule.
For LoRA, the dropout rate is set to $0.1$. We use a micro-batch size of 9 with 2 gradient accumulation steps per GPU.
The maximum sequence length is 3,072 tokens.

\begin{figure*}
  \centering
  \input{figures/annotation_prompts/llm_caption}
  \caption{
    \textbf{Prompt for generating the final natural language description.}
    This prompt is used with the 'minimal JSON' to generate the final natural language description of the modeling process.
  }
  \label{fig:annotation_llm}
\end{figure*}

\end{document}